\documentclass[]{openmoss}

\usepackage{helvet}
\usepackage{booktabs}
\usepackage{graphicx}
\usepackage{multirow}
\usepackage{tabularx}
\usepackage{array}
\usepackage{makecell}
\usepackage[table]{xcolor}
\usepackage{enumitem}
\usepackage{pifont}
\usepackage{amssymb}
\usepackage{amsmath}
\usepackage{adjustbox}
\usepackage{caption}
\usepackage{subcaption}
\usepackage{float}
\usepackage{url}
\usepackage{microtype}
\usepackage{nicefrac}
\usepackage{siunitx}
\usepackage{textcomp}

\definecolor{groupgray}{gray}{0.90}
\newcommand{\groupheader}[1]{%
  \addlinespace[2pt]
  \rowcolor{groupgray}
  \multicolumn{3}{c}{\textbf{#1}}\\
  \addlinespace[1pt]
}
\definecolor{oodteal}{HTML}{004B49}
\newcommand{\oodgain}[1]{\textcolor{oodteal}{\scriptsize $\uparrow$#1\%}}

\definecolor{gaincolor}{HTML}{178F86}
\definecolor{dropcolor}{HTML}{A66D61}
\newcommand{\gain}[1]{\,{\scriptsize\textcolor{gaincolor}{$\uparrow$#1\%}}}
\newcommand{\drop}[1]{\,{\scriptsize\textcolor{dropcolor}{$\downarrow$#1\%}}}

\title{Two Bridges, One Pathway: From VLMs to Generalizable VLAs with Embodied Trajectory-Coupled Data}

\author{
Linqi Yin$^{1,2,*}$ \hspace{.3em}
Shiduo Zhang$^{1,2,*,\ddagger}$ \hspace{.3em}
Shenling Qiu$^{1,*}$ \hspace{.3em}
Chenxin Li$^{1}$ \hspace{.3em}
Zhaoyang Fu$^{1}$ \hspace{.3em}
Lei Xiao$^{2}$ \\
Xiang Wang$^{2}$ \quad
Chenchen Yang$^{1,2}$ \quad
Zhe Xu$^{1,2}$ \quad
Pengfang Qian$^{1,2}$\\
Jingjing Gong$^{2}$ \quad
Xipeng Qiu$^{1,2}$ \quad
Xuanjing Huang$^{1}$ \quad
Yu-Gang Jiang$^{1,\dagger}$ \\
[1ex]
{\small
$^{1}$Fudan University \quad
$^{2}$Shanghai Innovation Institute \quad
}\\
{\small
$^{*}$Equal Contribution \quad
$^{\ddagger}$Project Lead \quad
$^{\dagger}$Corresponding Author
}
}

\abstract{
Vision-language models (VLMs) are powerful general-purpose reasoners, yet converting them into robot control policies (VLAs) is surprisingly difficult.
The root cause is a two-fold gap: VLMs are trained on internet-scale images with language-understanding objectives, while VLAs must perceive robot scenes and predict motor actions.
Fine-tuning a VLM directly on robot action data forces the model to cross both gaps at once---the learning curve is steep and the rich generalizations learned during pretraining tend to degrade rather than transfer.
We argue that this gap can be bridged gradually with the right intermediate data.
We introduce \emph{embodied trajectory-coupled (ETC) data}---vision-language supervision derived from the same robot scenes and trajectories used for action learning.
Because ETC data shares the visual context of robot operation while retaining familiar language-understanding objectives, it provides a natural stepping stone between VLM pretraining and VLA fine-tuning.
Building on this, we design a three-stage training recipe.
\textit{Distribution Bridging} first adapts the VLM to embodied visual-language semantics.
\textit{Objective Bridging} then gradually shifts the model toward action prediction while preserving the acquired representations.
\textit{Retentive Adaptation} finally specializes the policy to the target deployment domain.
We further show that mixing task-relevant out-of-distribution ETC data with a small amount of action data enables the model to generalize to novel visual-language conditions without requiring additional robot demonstrations.
Simulation and real-robot experiments confirm that this gradual bridging strategy is the key to transferring VLM generalization into robust, deployable robot policies.

\par\vspace{1.0\baselineskip}
\noindent\textbf{Homepage}: \url{https://linqiy.github.io/etc/}
}

\begin{document}
\maketitle

\section{Introduction}

\begin{figure*}[t]
  \centering
  \includegraphics[width=1\textwidth]{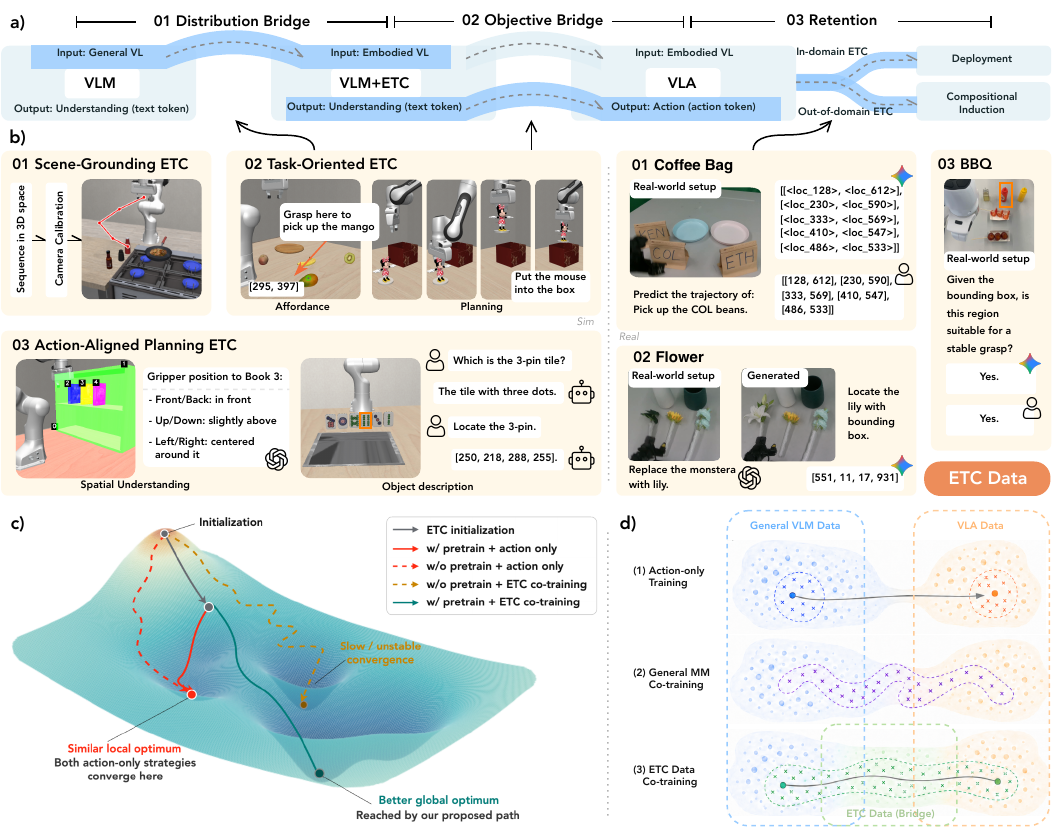}
    \caption{\textbf{Overview of the ETC-guided VLM-to-VLA paradigm.}
    (a) Three-stage ETC-based adaptation pathway: Distribution Bridging, Objective Bridging, and Retentive Adaptation, under a shared next-token prediction objective.
    (b) ETC examples and the offline construction pipeline.
    (c) A staged route from VLM initialization toward a generalizable VLA policy, compared with direct action-only adaptation and generic multimodal co-training.
    (d) ETC data bridges general vision-language data and robot action data, sharing embodied scene semantics with the latter while retaining vision-language supervision.}
  \label{fig:main}
  % \vspace{-1.2em}
\end{figure*}

Building a generalizable vision-language-action (VLA) policy from a pretrained vision-language model (VLM) requires crossing two distinct mismatches at the same time. On the \emph{input} axis, the VLM is trained on broad web-scale visual-language data, while a manipulation policy must operate on embodied scenes saturated with robot-specific viewpoints, object configurations, and interaction geometry. On the \emph{objective} axis, the VLM is optimized for next-token visual-language understanding, while the policy must generate executable actions. These two mismatches are orthogonal: closing one does not automatically close the other.

Prior work has sought to narrow this VLM-to-VLA gap through \emph{embodied reasoning enhancement}~\citep{gemini_robotics, embodiedmidtrain, zhai2025ignitingvlmsembodiedspace, xiaomi_robotics_0} before policy training and \emph{multimodal co-training} during VLA adaptation~\citep{zitkovich2023rt, pi05,lin2026systematic,chatvla}. These approaches aim to inherit common sense and visual-language understanding from VLMs ~\citep{driess2023palm, zitkovich2023rt}, enabling stronger embodied scene understanding ~\citep{robobrain, gemini_robotics} and more generalizable policies ~\citep{pi05, qu2026eo1openunifiedembodied}. While the former aims to equip the model with stronger embodied understanding, its value as a VLA initialization has not been systematically established~\citep{xiaomi_robotics_0, gr3, galaxea_g0}, and the acquired capability may still degrade once subsequent training switches to action-only supervision~\citep{knowledge_insulating_vla, wu2026pragmatic}. The latter \textit{co-training} strategy is often designed to retain broad VLM knowledge through generic multimodal data, but does not explicitly extend that knowledge toward robot scenes and manipulation tasks. More broadly, it remains unclear what kinds of multimodal supervision can most effectively benefit VLA policies, and at which adaptation stages they should be introduced. We therefore investigate three questions: (1) whether embodied multimodal supervision provides a stronger initialization for VLA formation? (2) what kind of multimodal data can most effectively bridge the dual gap between VLMs and VLAs? and (3) how to design a training recipe that expands the policy’s capability boundary while retaining as much pretrained knowledge as possible?

Our central hypothesis is that the dual VLM-to-VLA mismatch can be bridged through a data-centric intermediate supervision, which we call \emph{embodied trajectory-coupled (ETC) data}. ETC is pair-wise vision-language supervision derived from the same embodied scenes and trajectories as robot action data, shown in Figure \ref{fig:main}b. It retains the VLM’s next-token prediction objective while matching the embodied content of VLA training. As illustrated in Figure~\ref{fig:main}d, this dual alignment enables distribution alignment without action learning, followed by action grounding on an already aligned embodied representation. In this way, ETC separates distribution alignment from action grounding. We further develop a scalable offline pipeline that automatically derives ETC supervision from existing robot trajectories without additional human annotation.

Building on this insight, we organize a three-stage training paradigm under a shared next-token prediction objective (Figure~\ref{fig:main}a).  \textbf{(i) Distribution Bridging} fine-tunes the VLM backbone on ETC data, transporting the input distribution from common visual-language data onto the embodied manifold while leaving the objective unchanged. 
\textbf{(ii) Objective Bridging} introduces action-token prediction on this aligned representation while continuing ETC co-training from the same scenes and trajectories, preventing action learning from eroding the embodied alignment established in the first stage. 
% introduces the action objective on the now-aligned distribution while retaining ETC supervision as a knowledge-preservation signal: continued ETC co-training, drawn from the same scenes as the action data, keeps the backbone anchored to the embodied manifold so that learning to act does not erode the embodied understanding acquired in Distribution Bridging. 
\textbf{(iii) Retentive Adaptation} further adapts the policy to deployment domains with continued ETC supervision, maintaining reusable embodied competence while extending the policy’s capability boundary. 
Figure~\ref{fig:main}a and Figure~\ref{fig:main}c conceptually illustrate these two roles, respectively: ETC bridges general vision-language data and robot action data, while the three-stage pathway is designed as a smoother optimization route toward generalizable VLA policies than direct adaptation.

Beyond retaining embodied competence, we find that the two stages jointly trained with ETC and action data can support a new capability, which we call \emph{compositional induction}. By introducing task-relevant ETC supervision for unseen visual-language conditions, while pairing it with only limited in-distribution action data, the policy learns to associate new embodied conditions with already acquired action behaviors. This enables generalization to held-out visual-language compositions without collecting action demonstrations for those compositions.

Our contributions are three-fold:
\begin{itemize}[leftmargin=1.4em]
\item We introduce \emph{Embodied Trajectory-Coupled (ETC) data}, a data-centric bridge between VLMs and VLAs that is objective-aligned with VLM training and distribution-aligned with robot action data, together with an offline pipeline for deriving ETC supervision from existing trajectories.
\item We develop a three-stage ETC-based adaptation pathway—\textbf{Distribution Bridging}, \textbf{Objective Bridging}, and \textbf{Retentive Adaptation}—under a shared next-token prediction objective, progressively aligning embodied understanding, action generation, and downstream adaptation.
\item We identify \emph{compositional induction}: task-relevant out-of-distribution ETC, paired with limited in-distribution action data, enables policies to generalize to held-out visual-language compositions without action demonstrations for those conditions.
\end{itemize}

\section{Related Work}

\noindent\textbf{Knowledge Transfer and Preservation in Vision-Language-Action Models.} Adapting VLMs into generalizable VLA policies requires transferring multimodal knowledge into action learning while preserving the semantic and reasoning capabilities acquired during pretraining~\citep{pi0,gr00t_n1,liu2025faster}.
Recent work explores multimodal co-training and embodied adaptation to improve generalization and retain visiolinguistic competence~\citep{lin2026systematic,zhang2026vlm4vlarevisitingvisionlanguagemodelsvisionlanguageaction, vlm2vla, internvla_a1}.
% A growing body of work has pursued this goal empirically: joint vision-language and robot-action co-training has yielded promising open-world generalization~\citep{pi05,chatvla,chatvla2,gr3,agibot_world,xiaomi_robotics_0,internvla_a1,galaxea_g0}, and recent studies systematically examine which data modalities and training strategies best retain visiolinguistic competence during robot learning~\citep{lin2026systematic,zhang2026vlm4vlarevisitingvisionlanguagemodelsvisionlanguageaction,vlm2vla,contrastive_vla}. 
However, two gaps remain. First, generic multimodal co-training relies on supervision whose distribution is distant from embodied action data, making it an indirect bridge from VLM knowledge to manipulation-relevant behavior~\citep{lin2026systematic, robo2vlm, robobrain}. 
% Yet two recurring failure modes remain unresolved. First, co-training with general multimodal data~\citep{a_okvqa,latex_ocr} suffers from a fundamental distribution mismatch: VLM and VLA data occupy largely disjoint representation spaces~\citep{embodiedmidtrain,lin2026systematic}, forcing the model to balance two incompatible objectives rather than follow a coherent learning curriculum~\citep{zhai2025ignitingvlmsembodiedspace,chatvla,chatvla2,xiaomi_robotics_0}. 
Second, off-the-shelf VLMs provide limited embodied initialization, while embodied mid-training, although beneficial, does not by itself ensure that acquired embodied capabilities survive subsequent action learning~\citep{embodiedmidtrain,zhang2026vlm4vlarevisitingvisionlanguagemodelsvisionlanguageaction,galaxea_g0}. 
% Second, initializing directly from off-the-shelf VLMs offers a weak starting point, since general-purpose pretraining on captioning and VQA produces representations poorly aligned with the spatial and physical reasoning that action generation demands~\citep{embodiedmidtrain,zhang2026vlm4vlarevisitingvisionlanguagemodelsvisionlanguageaction}; embodied mid-training can mitigate the initialization gap~\citep{embodiedmidtrain,galaxea_g0}, but leaves open how knowledge is sustained \emph{throughout} action learning. 
These limitations motivate a multi-stage adaptation framework that explicitly studies how embodied supervision supports knowledge transfer, retention, and downstream capability expansion across VLA training stages.
% Together, these gaps point to a missing piece: a principled multi-stage adaptation framework that explicitly addresses knowledge transfer across initialization, action learning, and post-training, and that mechanistically characterizes the role of embodied supervision at each stage.

\noindent\textbf{Embodied Understanding and Reasoning in Vision-Language Models.} 
A parallel line of work equips VLMs with manipulation-relevant understanding through embodied multimodal supervision. Before policy training, robotics-oriented VQA~\citep{robovqa,vlm2vla}, spatial grounding~\citep{embspatial,roborefer,robopoint}, and embodied foundation models introduce priors over planning~\citep{molmoact,robopoint}, affordances~\citep{robo2vlm,gemini_robotics}, interaction geometry~\citep{robobrain,robobrain2}, and temporal reasoning~\citep{mimo_embodied,zhai2025ignitingvlmsembodiedspace}.
Yet these studies largely treat embodied supervision as either an initialization source or a task-specific auxiliary signal. Our work instead studies trajectory-coupled supervision as a shared bridge across adaptation stages, and analyzes its role in inducing compositional generalization to unseen visual-language conditions.
\section{Method}
\label{sec:method}

% \noindent\textbf{Architecture.} We build on $\pi_0$-FAST~\citep{pi0,pertsch2025fastefficientactiontokenization}, which discretizes continuous actions into FAST tokens atop a pretrained VLM backbone. This unifies ETC and action supervision under a shared next-token prediction objective.

\subsection{Embodied Trajectory-Coupled Data}
\label{subsec:etc_data}
% TODOs: 引主图
We define \emph{Embodied Trajectory-Coupled (ETC) data} as vision-language supervision derived from the same embodied scenes and trajectories as robot action data. ETC retains the VLM's next-token prediction objective while matching the VLA embodied distribution. This dual alignment enables distribution bridging before action learning and action grounding over aligned representations.

We instantiate ETC as VQA-style supervision in three categories of decreasing abstraction, forming a graded spectrum from common-sense understanding to motor behaviors (examples in Figure~\ref{fig:main}b). \noindent\textbf{Scene-Grounding ETC} captures task-agnostic scene perception: spatial queries probe geometric relationships (relative direction to the end effector, cross-view object correspondence), and object-description queries identify, localize, or describe manipulation-relevant attributes of a target object. \noindent\textbf{Task-Oriented ETC} is instruction-conditioned: affordance queries predict optimal grasp points or validate whether a region affords a stable grasp, and task-planning queries decompose an instruction into ordered sub-actions, recognize the current phase, or anticipate the next action. \noindent\textbf{Action-Aligned Planning ETC} sits closest to motion, forecasting the gripper's future trajectory as image-plane key points for a given sub-task—the final stepping stone before action tokens. We derive ETC offline from existing robot trajectories, without additional human annotation: geometric labels are computed deterministically via camera calibration (or by unifying actions into the camera frame when calibration is unavailable~\citep{xie2026unifyrobotactionscamera}); semantic and reasoning labels are generated by Gemini. The full pipeline and examples are in Appendix~\ref{sec:vqa_construction_pipeline}.

\subsection{Three-Stage ETC-based Adaptation}
\label{subsec:three_stage}
% 3stage 定义&具体怎么做
We adapt a pretrained VLM into a VLA through three stages, all sharing the same next-token prediction objective. The stages map directly onto the dual VLM-to-VLA mismatch: Stage~1 closes the input-distribution gap on the VLM objective; Stage~2 closes the objective gap on the already-aligned distribution; Stage~3 keeps both gaps closed while shifting to deployment scenes.

% TODOs: 这里要不要提到训练框架？
\noindent\textbf{Distribution Bridging (Stage~1).} The first stage operates purely on the VLM backbone and transports its input distribution from general visual-language data onto the embodied manifold, while leaving the next-token objective unchanged. We train the backbone on ETC data alone, spanning all three categories defined in Section~\ref{subsec:etc_data}, and unfreeze the ViT tower so that embodied supervision can align semantic visual features with action-relevant geometry.

\noindent\textbf{Objective Bridging (Stage~2).} The second stage introduces the action objective on top of the aligned distribution from Stage~1, co-training large-scale pretraining action data with ETC drawn from the same scenes and trajectories. ETC's role shifts from \emph{establishing} the embodied distribution to \emph{preserving} it, so that the action objective extends the model from understanding to generation without overwriting Stage~1's alignment.

\noindent\textbf{Retentive Adaptation (Stage~3).} The third stage adapts the policy to downstream deployment scenes by retaining the co-training scheme of Stage~2, now pairing action data with ETC constructed for the target scenes. The input distribution shifts to new objects, viewpoints, and layouts, but ETC's dual-alignment property carries forward: continued ETC supervision preserves reusable embodied competence, while the action objective extends the policy's capability boundary onto the target tasks. Stage~3 thus realizes \emph{capability expansion without knowledge erosion} as a training recipe.

\subsection{Compositional Induction via ETC Co-training}
\label{subsec:compositional_induce}
The two stages jointly trained with ETC and action data support \emph{compositional induction}. For target compositions without action demonstrations, we construct ETC covering these held-out conditions and mix it into the co-training stream. Sharing the same backbone and objective, the policy applies in-distribution skills to conditions seen only through ETC, generalizing without collecting actions for them. We validate this in simulation and on real robots.
\section{Experiments}

% benchmark & metrics (sr定义) & 真机
% baseline -> 和自己比
% Data配置
% 
% cotrain的实现细节一句话交代

\subsection{Experiment Setup}
\label{sec:experiment_setup}

% TODO: 可能还是要修改一下 & 简化一下
\noindent\textbf{Training and Evaluation.} We train all the models from Paligemma~\citep{beyer2024paligemma} and train VLA following Pifast~\citep{pertsch2025fastefficientactiontokenization} with a unified next-token prediction objective for both ETC and action supervision. All training details are provided in Appendix~\ref{sec:implementation_details}. We evaluate on three simulation benchmarks---LIBERO~\citep{libero}, SimplerEnv~\citep{simpler}, and VLABench~\citep{vlabench}---and a real-world WidowX platform with two tasks: \textit{Inserting Flower} and \textit{Placing Coffee Bag}, both instantiated on a BridgeV2-pretrained Pifast~\citep{walke2023bridgedata}. These benchmarks vary in how much they require embodied understanding beyond action execution alone, and we report success rate under each benchmark's standard protocol. Each benchmark is paired with domain-matched ETC supervision: ShareRobot~\citep{robobrain} and VLA-OS~\citep{vlaos} are adopted for SimplerEnv and LIBERO, respectively, while VLABenchVQA and real-world ETC are constructed in-house. Full statistics are in Appendix~\ref{sec:data}.

% \noindent\textbf{Controlled Comparison.} All variants follow the same PaliGemma VLM architecture and $\pi_0$-FAST architecture and three-stage framework (Section~\ref{sec:method}), differing only in the training configuration under study and defined in the corresponding analyses.

\noindent\textbf{Embodied Understanding Benchmarks.} We also probe the VLM backbone's multimodal understanding on a held-out set of ETC and general multimodal data, evaluated on both the initialized VLM and the midtrain VLA backbone to track whether embodied understanding is acquired and preserved across stages. Full evaluation details are provided in Appendix~\ref{sec:emm_bench_details}.

\subsection{Analyzing the Three-Stage ETC-based Adaptation Pathway}

\begin{table}[h]
  \centering
  \caption{\textbf{Comparison of Stage~2 supervision strategies across different Distribution Bridging settings.} Rows are grouped by Stage~2 supervision; \textit{w/ Distribution Bridging} annotations report relative changes over the corresponding \textit{w/o Distribution Bridging} row.}
  \label{tab:vla_sr_cotrain}
  \scriptsize
  \setlength{\tabcolsep}{2.8pt}
  \resizebox{\textwidth}{!}{%
  \begin{tabular}{llcccccccc}
    \toprule
    & & & & \multicolumn{3}{c}{VLABench T1} & \multicolumn{3}{c}{VLABench T2} \\
    \cmidrule(lr){5-7} \cmidrule(lr){8-10}
    Stage~2 signal & Backbone / Stage~1 init. & Simpler & LIBERO
      & SR & PS & IS & SR & PS & IS \\
    \midrule

\multirow{3}{*}{Action only}
  & SigLIP--Gemma
  & 0.385
  & \textbf{0.915}
  & -- & -- & -- & -- & -- & -- \\
  & w/o Distribution Bridging
  & 0.792
  & 0.894
  & 0.372
  & 0.482
  & 0.508
  & 0.236
  & 0.312
  & 0.306 \\
  & w/ Distribution Bridging
  & \underline{0.812}\gain{2.5}
  & \underline{0.907}\gain{1.5}
  & 0.443\gain{19.1}
  & 0.531\gain{10.2}
  & 0.537\gain{5.7}
  & 0.283\gain{19.9}
  & 0.351\gain{12.5}
  & 0.365\gain{19.3} \\

\addlinespace[3pt]

\rowcolor{gray!12}
  & w/o Distribution Bridging
  & 0.677
  & 0.816
  & 0.381
  & 0.490
  & 0.518
  & 0.241
  & 0.328
  & 0.353 \\
\rowcolor{gray!12}
\multirow{-2}{*}{$+$ MM}
  & w/ Distribution Bridging
  & 0.729\gain{7.7}
  & 0.833\gain{2.1}
  & 0.610\gain{60.1}
  & 0.726\gain{48.2}
  & 0.740\gain{42.9}
  & \underline{0.393}\gain{63.1}
  & \underline{0.464}\gain{41.5}
  & \textbf{0.473}\gain{34.0} \\

\addlinespace[3pt]

\multirow{2}{*}{$+$ ETC}
  & w/o Distribution Bridging
  & 0.802
  & 0.869
  & 0.424
  & 0.525
  & 0.519
  & 0.226
  & 0.316
  & 0.364 \\
  & w/ Distribution Bridging
  & \textbf{0.844}\gain{5.2}
  & 0.868\drop{0.1}
  & \textbf{0.648}\gain{52.8}
  & \textbf{0.773}\gain{47.2}
  & \textbf{0.778}\gain{49.9}
  & \textbf{0.399}\gain{76.5}
  & \textbf{0.478}\gain{51.3}
  & \underline{0.471}\gain{29.4} \\

\rowcolor{gray!12}
  & w/o Distribution Bridging
  & 0.708
  & 0.825
  & 0.412
  & 0.498
  & 0.493
  & 0.260
  & 0.334
  & 0.327 \\
\rowcolor{gray!12}
\multirow{-2}{*}{$+$ MM+ETC}
  & w/ Distribution Bridging
  & 0.729\gain{3.0}
  & 0.901\gain{9.2}
  & \underline{0.611}\gain{48.3}
  & \underline{0.727}\gain{46.0}
  & \underline{0.744}\gain{50.9}
  & 0.369\gain{41.9}
  & 0.449\gain{34.4}
  & 0.463\gain{41.6} \\

    \bottomrule
  \end{tabular}%
  }
\end{table}

\subsubsection{Distribution Bridging: Aligning the VLM to the Embodied Manifold}

% VQA
% \begin{table}[t]
%   \centering
%   \caption{Performance on held-out, domain-matched manipulation-relevant VQA
%   benchmarks before action training. Each EMM initialization is evaluated on
%   the benchmark corresponding to its Stage~1 embodied supervision domain.
%   Detailed per-type results are provided in
%   Appendix~\ref{sec:emm_bench_details}.}
%   \label{tab:emm_vlm_bench}
%   \small
%   \setlength{\tabcolsep}{5pt}
%   \begin{tabular}{lcccc}
%     \toprule
%     Stage~1 domain & Evaluation benchmark & Raw init. & EMM init. & $\Delta$ \\
%     \midrule
%     Bridge   & ShareRobot    &  5.44 &  \textbf{8.88} & +3.44  \\
%     LIBERO   & VLA-OS        & 12.21 & \textbf{41.21} & +29.00 \\
%     VLABench & VLABenchVQA   &  9.55 & \textbf{56.38} & +46.83 \\
%     DROID    & Robo2VLM-VQA  & 15.15 & \textbf{69.29} & +54.14 \\
%     \bottomrule
%   \end{tabular}
% \end{table}

% different init
% \begin{table}[t]
%   \centering
%   \caption{Bridge success rates under different backbone initialization settings.
%   For EMM initialization, we compare freezing versus updating the ViT tower
%   during pretraining.}
%   \label{tab:bridge_init_ablation}
%   \small
%   \setlength{\tabcolsep}{8pt}
%   \begin{tabular}{lc}
%     \toprule
%     Backbone initialization & Bridge SR \\
%     \midrule
%     SigLIP                         & 0.385 \\
%     PaliGemma w/o EMM init.        & 0.792 \\
%     PaliGemma w/ EMM init. (freeze ViT) & 0.673 \\
%     PaliGemma w/ EMM init.         & \textbf{0.812} \\
%     \bottomrule
%   \end{tabular}
% \end{table}

\begin{figure*}[h]
  \centering
  \includegraphics[width=1\textwidth]{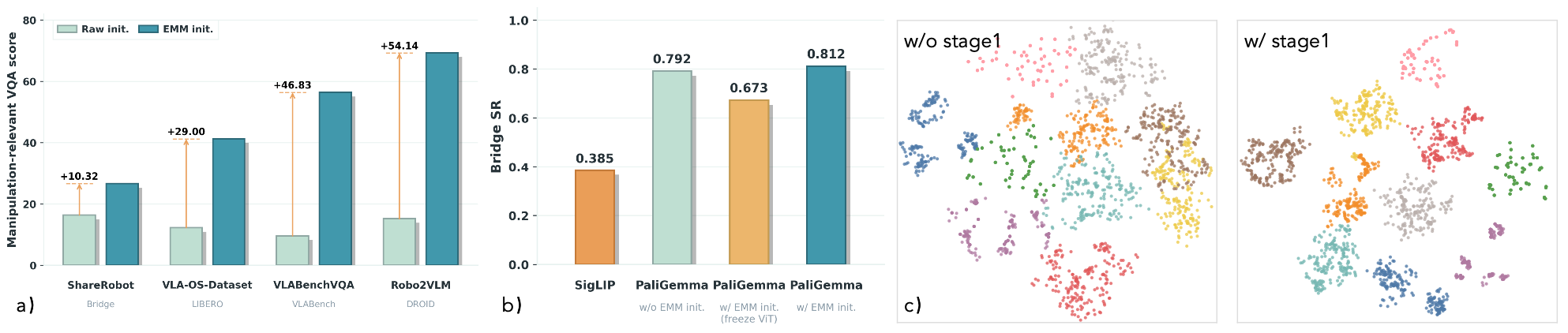}
    \caption{\textbf{Distribution Bridging aligns the VLM to the embodied manifold.}
    (a) ETC supervision improves performance on embodied understanding benchmarks.
    (b) Comparison of different Stage 1 strategies.
    (c) t-SNE comparison on 10 VLABench tasks: w/ vs.\ w/o Stage~1.}
  \label{fig:distribution_bridging}
\end{figure*}

Effective Distribution Bridging requires adapting embodied representations across both vision and language. We show that this stage provides a stronger basis for subsequent action learning, with its benefit arising primarily from joint vision-language adaptation rather than language-only tuning.

\noindent\textbf{Does Distribution Bridging Provide a Better Starting Point for Action Learning?} 
The Distribution-Bridged VLM backbone consistently outperforms the raw PaliGemma backbone across nearly all Stage~2 settings, with the largest gains on VLABench (Table~\ref{tab:vla_sr_cotrain}). Under action-only training, which isolates the effect of initialization, Distribution Bridging improves SR by $+2.0$ points on SimplerEnv, $+1.3$ on LIBERO, $+7.1$ on VLABench T1, and $+4.7$ on T2. The Distribution-Bridged VLM achieves higher ETC-bench scores than the raw VLM (Figure~\ref{fig:distribution_bridging}a), and the t-SNE visualization (Figure~\ref{fig:distribution_bridging}c) further shows that the Distribution-Bridged VLM separates task-conditioned embodied inputs that the raw VLM mixes together, suggesting embodied task structure has been injected into the VLM representation as a stronger initialization.

\noindent\textbf{Why Is Distribution Bridging Needed beyond General VLM Pretraining?}
Figure~\ref{fig:distribution_bridging}b supports our claim that general VLM representations remain misaligned with embodied manipulation scenes. With identical action-only Stage~2 training, raw PaliGemma clearly outperforms SigLIP–Gemma ($0.792$ vs.\ $0.385$), showing the value of joint vision–language pretraining. Yet ETC-based Distribution Bridging further raises SR to $0.812$, indicating that general VLM pretraining alone is insufficient for embodied alignment. Freezing the ViT during bridging instead reduces SR to $0.673$, further suggesting that the remaining gap requires visual adaptation, not just language-side tuning.
% Figure~\ref{fig:distribution_bridging}b shows that Distribution Bridging requires joint visual--language adaptation. Under the same action-only Stage~2 training, the raw PaliGemma backbone substantially outperforms the SigLIP--Gemma baseline that lacks joint vision--language pretraining ($0.792$ vs.\ $0.385$), while freezing the ViT during Distribution Bridging reduces SR to $0.673$, suggesting an embodied-manifold mismatch between the adapted language backbone and fixed visual features. Updating both components instead achieves the best SR of $0.812$.

% =================================================

\subsubsection{Objective Bridging: Preserving Embodied Alignment under Action Learning}
\label{subsec:objective_bridging_subsection}

% \begin{figure*}[h]
%   \centering
%   \includegraphics[width=1\textwidth]{figures/midtrain_v3.png}
%   \caption{
%   % TODO
%   a) midtrain cka compare
%   b) VLABench performance under different Stage~2 EMM ablations. Best results for each metric are shown in bold.
%   }
%   \label{fig:cotrain-cka}
% \end{figure*}

\begin{figure*}[t]
  \centering

  % Main contents
  \begin{minipage}[t]{0.45\textwidth}
    \centering
    \vspace{0pt}
    \includegraphics[width=\linewidth]{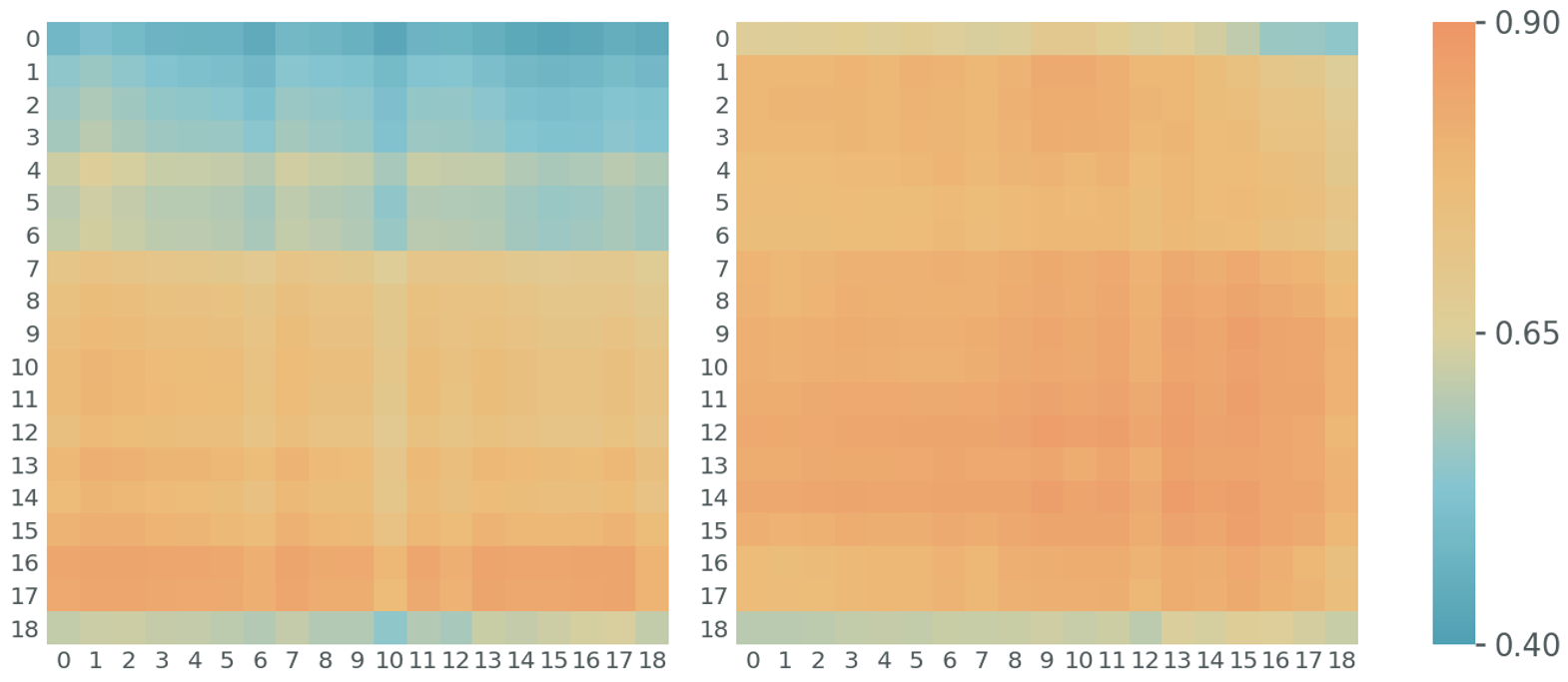}
    \vspace{1pt}
    \resizebox{\linewidth}{!}{\scriptsize Orange = more similar layer representations; cyan = less similar.}
  \end{minipage}
  \hfill
  \begin{minipage}[t]{0.54\textwidth}
    \centering
    \vspace{0.8em}
    \scriptsize
    \setlength{\tabcolsep}{2.5pt}
    \begin{tabular}{lcccccc}
      \toprule
      \multirow{2}{*}{Stage~2 supervision}
        & \multicolumn{3}{c}{Track~1}
        & \multicolumn{3}{c}{Track~2} \\
      \cmidrule(lr){2-4}
      \cmidrule(lr){5-7}
        & SR & PS & IS & SR & PS & IS \\
      \midrule
      Action only
        & 0.443 & 0.531 & 0.537
        & 0.283 & 0.351 & 0.365 \\
      $+$ Scene-grounding ETC
        & 0.562 & 0.696 & 0.736
        & 0.407 & 0.481 & 0.473 \\
      $+$ Task-oriented ETC
        & 0.593 & 0.719 & 0.748
        & 0.314 & 0.423 & \underline{0.507}\\
      $+$ Action Planning ETC
        & \underline{0.621}& \underline{0.744}& \underline{0.771}& \textbf{0.438} & \textbf{0.499} & 0.482 \\
      $+$ Full ETC
        & \textbf{0.648} & \textbf{0.773} & \textbf{0.778}
        & \underline{0.414}& \underline{0.498}& \textbf{0.511} \\
      \bottomrule
    \end{tabular}
  \end{minipage}

  % Aligned subfigure labels
  \vspace{3pt}

  \begin{minipage}[t]{0.45\textwidth}
    \centering
    \textbf{(a)}
  \end{minipage}
  \hfill
  \begin{minipage}[t]{0.54\textwidth}
    \centering
    \textbf{(b)}
  \end{minipage}
    \caption{\textbf{Objective Bridging: ETC co-training preserves embodied alignment under action learning.}
    (a) CKA between the Stage~2 backbone and the Distribution-Bridged VLM. Left: action-only Objective Bridging; right: ETC co-training.
    (b) VLABench performance under different ETC category ablations during Objective Bridging, on Track~1 (in-domain) and Track~2 (cross-category).}
    \label{fig:objective_bridging}
\end{figure*}

\noindent\textbf{Is ETC Co-training Still Needed after Distribution Bridging?}
We focus on SimplerEnv and VLABench, since LIBERO is less diagnostic for embodied preservation: even SigLIP--Gemma reaches $0.915$ SR with action-only training (Table~\ref{tab:vla_sr_cotrain}). We also exclude SigLIP--Gemma from subsequent analyses, as its lack of joint vision--language pretraining makes it an inadequate starting point for evaluating ETC's preservation effect. With the same Distribution-Bridged backbone, ETC co-training improves SR by $+0.032$ on SimplerEnv, with larger gains on VLABench T1 ($+0.205$) and T2 ($+0.116$). The CKA analysis (Figure~\ref{fig:objective_bridging}a) shows that action-only Objective Bridging drifts farther away from the Distribution-Bridged VLM representation, while ETC co-training keeps the backbone substantially closer to it. This indicates that ETC co-training mitigates representational forgetting and preserves the embodied vision-language alignment acquired through Distribution Bridging.

\noindent\textbf{Why Is ETC More Effective than General Multimodal Supervision?}
We ablate the three ETC categories under the same Distribution-Bridged backbone and identical action data, isolating the contribution of each signal(Fig.~\ref{fig:objective_bridging}b). Every category improves over action-only training across all metrics, but their strengths mirror their content. Action-aligned Planning ETC, the signal most tightly coupled to motion, gives the strongest execution, achieving the best single-category SR and PS on both tracks (e.g., $0.621$ SR on Track~1 and $0.438$ on Track~2). Task-oriented ETC, which supervises intention and task decomposition, instead excels at understanding, attaining the highest Intention Score on the cross-category Track~2 ($0.507$). Scene-grounding ETC improves all metrics as a broad perceptual foundation but does not dominate any single one. Combining all three (Full ETC) is best across Track~1 and yields the highest Track~2 Intention Score ($0.511$), though action-aligned grounding remains strongest on Track~2 execution---indicating that the three signals are largely complementary, each contributing the capability its content most directly supports.

% \begin{table}[h]
%     \centering
%     \caption{VLABench Track~1 in-distribution and Track~2 cross-
%     category
%     generalization under different Stage~2 EMM supervision
%     signals. All
%     conditions use PaliGemma-EMM initialization with identical
%     in-domain action
%     training data. Best results in each column are shown in
%     \textbf{bold}.}
%     \label{tab:vlabench_emm_ablation}
%     \small
%     \setlength{\tabcolsep}{4pt}
%     \begin{tabular}{lcccccc}
%       \toprule
%       \multirow{2}{*}{Stage~2 supervision}
%         & \multicolumn{3}{c}{Track~1}
%         & \multicolumn{3}{c}{Track~2} \\
%       \cmidrule(lr){2-4}
%       \cmidrule(lr){5-7}
%         & SR & PS & IS & SR & PS & IS \\
%       \midrule
%       Action only
%         & 0.443 & 0.531 & 0.537
%         & 0.283 & 0.351 & 0.365 \\
%       $+$ Scene-grounding EMM
%         & 0.562 & 0.696 & 0.736
%         & 0.407 & 0.481 & 0.473 \\
%       $+$ Task-oriented EMM
%         & 0.593 & 0.719 & 0.748
%         & 0.314 & 0.423 & 0.507 \\
%       $+$ Action-aligned Planning EMM
%         & 0.621 & 0.744 & 0.771
%         & \textbf{0.438} & \textbf{0.499} & 0.482 \\
%       $+$ Full EMM
%         & \textbf{0.648} & \textbf{0.773} & \textbf{0.778}
%         & 0.414 & 0.498 & \textbf{0.511} \\
%       \bottomrule
%     \end{tabular}
% \end{table}

% TODO: 这里可以把task 2上多挖掘一点
\noindent\textbf{Which Category of ETC benefits More?} 
We ablate three ETC categories under the same Distribution-Bridged backbone and action data (Figure~\ref{fig:objective_bridging}b). All outperform action-only training, with gains reflecting their supervision content: Action-aligned Planning ETC leads execution on both Track~1 and the cross-category Track~2, where its motion-coupled geometry transfers across task compositions; Task-oriented ETC instead leads understanding, attaining the highest Track~2 Intention Score ($0.507$); Scene-grounding ETC provides a broad perceptual foundation, lifting all metrics without dominating any. Full ETC performs best on Track~1, indicating that the three signals are complementary.

% \begin{table}[h]
%   \centering
%   \caption{VLABench Track~2 cross-category generalization under different Stage~2 EMM supervision and data-steering settings. All conditions use PaliGemma-EMM initialization with identical in-domain action training data. \emph{Full EMM} uses in-domain EMM supervision, whereas \emph{OOD EMM Steer} pairs the same action data with OOD EMM supervision under the reported 0.94/0.06 setting. Best results in each column are shown in \textbf{bold}.}
%   \label{tab:track2_ood}
%   \small
%   \setlength{\tabcolsep}{5pt}
%   \begin{tabular}{lccc}
%     \toprule
%     Stage~2 supervision & SR & PS & IS \\
%     \midrule
%     Action only
%       & 0.283 & 0.351 & 0.365 \\
%     \midrule
%     \multicolumn{4}{l}{\emph{In-domain EMM supervision}} \\
%     $+$ Understanding-oriented EMM
%       & 0.407 & 0.481 & 0.473 \\
%     $+$ Task-level reasoning EMM
%       & 0.314 & 0.423 & 0.507 \\
%     $+$ Action-aligned grounding EMM
%       & \textbf{0.438} & \textbf{0.499} & 0.482 \\
%     $+$ Full EMM
%       & 0.414 & 0.498 & 0.511 \\
%     \midrule
%     \multicolumn{4}{l}{\emph{OOD EMM data steering}} \\
%     $+$ OOD EMM Steer (0.94/0.06)
%       & 0.389 & 0.482 & \textbf{0.533} \\
%     \bottomrule
%   \end{tabular}
% \end{table}
 
\subsubsection{Retentive Adaptation: Extending Capability without Knowledge Erosion}
\begin{figure*}[h]
  \centering

  % Main contents
  \begin{minipage}[t]{0.54\textwidth}
    \centering
    \vspace{1.12em}
    \scriptsize
    \setlength{\tabcolsep}{2.5pt}
      \centering
      \small
      \setlength{\tabcolsep}{8pt}
      \begin{tabular}{cccc}
        \toprule
        Stage~1 & Stage~2 & Stage~3 & In-domain Score\\
        \midrule
        \ding{55} & \ding{55} & \ding{55} & 66.67 \\
        \ding{55} & \ding{55} & \ding{51} & 61.11 \\
        \ding{55} & \ding{51} & \ding{55} & 75.00 \\
        \ding{55} & \ding{51} & \ding{51} & \textbf{95.83} \\
        \ding{51} & \ding{55} & \ding{55} & 66.67 \\
        \ding{51} & \ding{55} & \ding{51} & 72.22 \\
        \ding{51} & \ding{51} & \ding{55} & \underline{91.67}\\
        \ding{51} & \ding{51} & \ding{51} & \textbf{95.83} \\
        \midrule
        \multicolumn{3}{c}{From scratch} & 63.88 \\
        \bottomrule
      \end{tabular}
  \end{minipage}
  \hfill
  \begin{minipage}[t]{0.45\textwidth}
    \centering
    \vspace{0pt}
    \includegraphics[width=\linewidth]{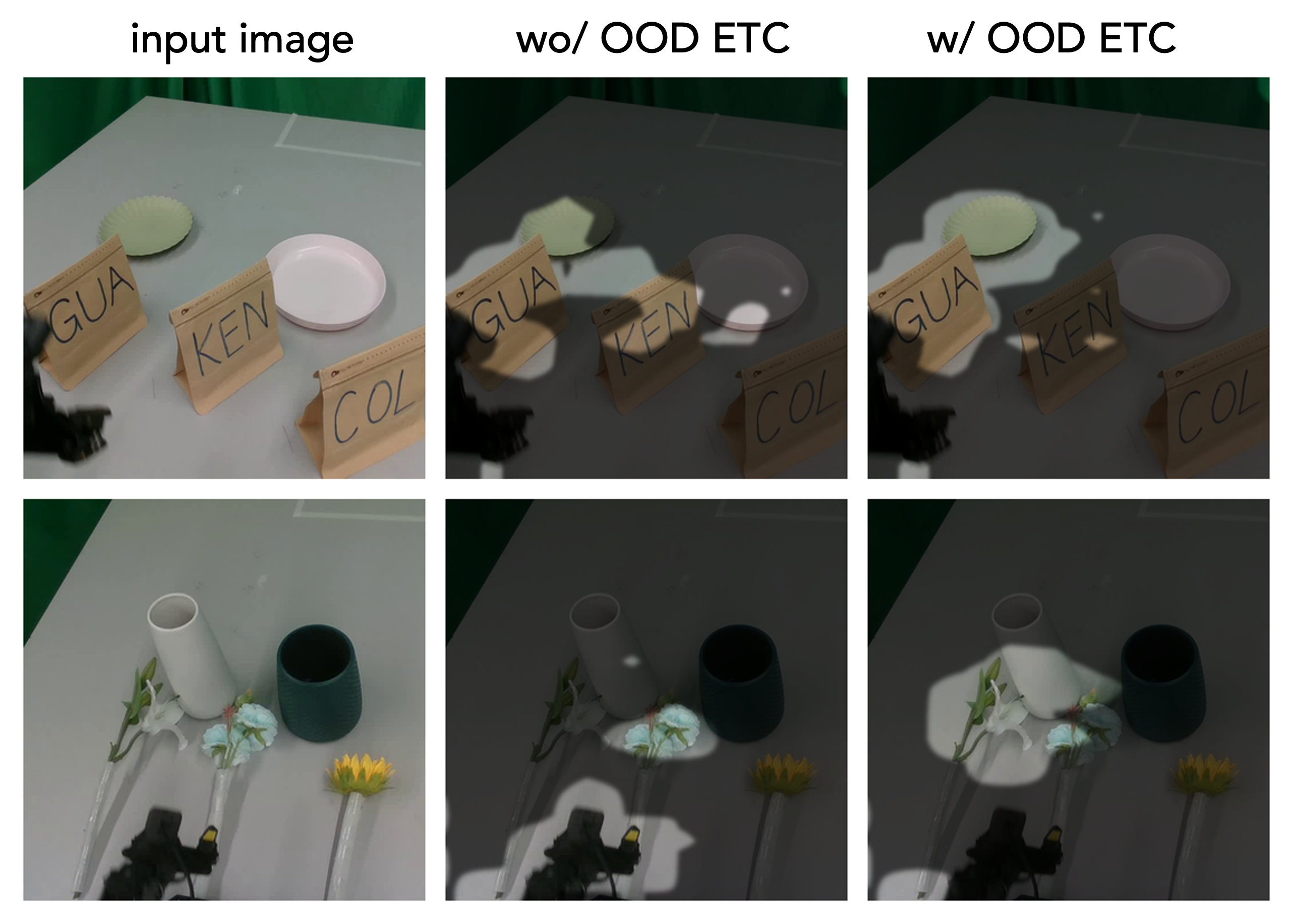}
  \end{minipage}

  % Aligned subfigure labels
  \vspace{3pt}

  \begin{minipage}[t]{0.45\textwidth}
    \centering
    \textbf{(a)}
  \end{minipage}
  \hfill
  \begin{minipage}[t]{0.54\textwidth}
    \centering
    \textbf{(b)}
  \end{minipage}
    \caption{\textbf{Retentive Adaptation: extending capability without knowledge erosion.} (a) Three-stage ETC adaptation ablation on the WidowX \textit{pick coffee bag into the plate} task. (b) Attention heatmaps under OOD targets. Rows show different real-robot examples. Columns show the original observation, the policy without OOD ETC, and the policy with OOD ETC.}
    \label{fig:retentive}
\end{figure*}

% 这里放真机实验的表格。简单叙述在真机上的结果和在仿真中的相似就可以；然后放上真机SR
The real-robot evaluation further validates \textbf{Retentive Adaptation} on a downstream deployment task (Figure~\ref{fig:retentive}a). Consistent with the simulation results, continuing ETC supervision through Stage 2 and Stage 3 yields the strongest real-robot performance, whereas introducing ETC only during Stage 3 is substantially less effective. This shows that Stage~3 benefits from adapting a policy with reusable embodied competence retained from the preceding stages.

% caption
% \textbf{Ablation of the three-stage ETC-based adaptation pathway} on a WidowX real-robot task. Each stage is independently toggled: Stage~1 \textbf{Distribution Bridging} (ETC pretraining), Stage~2 \textbf{Objective Bridging} (ETC co-training with action data), and Stage~3 \textbf{Retentive Adaptation} (ETC co-training during post-training). Action post-training is always performed. \textit{From scratch} denotes training without a pretrained VLM backbone. Scores are over 18 rollouts with partial credit. Best in \textbf{bold}.

% \begin{table}[h]
%   \centering
%     \caption{Three-stage ETC adaptation ablation on the WidowX pick coffee bag into the plate task}
%   \label{tab:realrobot_stage_ablation}
%   \small
%   \setlength{\tabcolsep}{8pt}
%   \begin{tabular}{cccc}
%     \toprule
%     Stage~1 & Stage~2 & Stage~3 & In-domain SR \\
%     \midrule
%     \ding{55} & \ding{55} & \ding{55} & 66.67 \\
%     \ding{55} & \ding{55} & \ding{51} & 61.11 \\
%     \ding{55} & \ding{51} & \ding{55} & 75.00 \\
%     \ding{55} & \ding{51} & \ding{51} & \textbf{95.83} \\
%     \ding{51} & \ding{55} & \ding{55} & 66.67 \\
%     \ding{51} & \ding{55} & \ding{51} & 72.22 \\
%     \ding{51} & \ding{51} & \ding{55} & 91.67 \\
%     \ding{51} & \ding{51} & \ding{51} & \textbf{95.83} \\
%     \midrule
%     \multicolumn{3}{c}{From scratch} & 63.88 \\
%     \bottomrule
%   \end{tabular}
% \end{table}

\subsection{Compositional Induction via Constructible OOD ETC}

Beyond preserving and adapting in-distribution knowledge, the two stages jointly trained with ETC and action data enable compositional induction: generalizing to held-out visual-language compositions without action demonstrations for them.

\subsubsection{OOD ETC Induces Compositional Generalization}
We first evaluate this effect on VLABench Track2 (Table\ref{tab:realrobot_compositional_induction}). We augment mid-training with OOD ETC at a 94{:}6 in-domain to out-of-distribution ratio while keeping all action demonstrations in-domain. Despite this small intervention, OOD ETC further improves Track2 PS and IS, showing that ETC supervision can associate unseen task conditions with already acquired action behaviors without requiring OOD action data.
% Section~\ref{subsec:objective_bridging_subsection}
% Section~4.2.2 shows that Distribution Bridging and Objective Bridging build a stronger generalization foundation, improving SR, PS, and IS on the cross-category Track~2. We now ask whether unseen conditions can be introduced through ETC alone. We augment mid-training with OOD ETC at a 94{:}6 in-domain-to-OOD ratio, while keeping all action demonstrations in-domain. This small intervention further improves Track~2 PS and IS, showing that OOD ETC associates unseen task conditions with already-acquired actions---without any OOD action data.

\begin{table}[t]
\centering
\caption{\textbf{Compositional induction via OOD ETC supervision} on VLABench Track~2 and two real-robot tasks. All variants share the same Distribution-Bridged and Objective-Bridged policy.}
\label{tab:realrobot_compositional_induction}
\small
\setlength{\tabcolsep}{5pt}
\begin{tabular}{llccc}
\toprule
\rowcolor{gray!15}
\multicolumn{5}{l}{\textbf{\textit{Simulation evaluation}}} \\
\addlinespace[2pt]
Benchmark & Stage~3 & SR & PS & IS \\
\midrule
\multirow{2}{*}{VLABench Track~2}
  & ID(in domain) ETC only
  & \textbf{0.399}
  & 0.478
  & 0.471 \\
  & ID ETC $+$ OOD ETC
  & 0.389
  & \textbf{0.482}\oodgain{0.8}
  & \textbf{0.533}\oodgain{13.2} \\
\addlinespace[4pt]
\rowcolor{gray!15}
\multicolumn{5}{l}{\textbf{\textit{Real-robot evaluation}}} \\
\addlinespace[2pt]
Task & Stage~3 & ID Score & \multicolumn{2}{c}{OOD Score} \\
\midrule
\multirow{4}{*}{Coffee bag}
  & Action only                     & \underline{91.67}          & \multicolumn{2}{c}{41.67}          \\
  & $+$ ID ETC               & \textbf{95.83} & \multicolumn{2}{c}{69.44}          \\
  & $+$ OOD ETC (Human)             & \underline{91.67}          & \multicolumn{2}{c}{\underline{80.56}}          \\
  & $+$ OOD ETC (Gemini)             & \underline{91.67}         & \multicolumn{2}{c}{\textbf{91.67}} \\
\midrule
\multirow{4}{*}{Flower}
  & Action only                     & \textbf{47.22} & \multicolumn{2}{c}{27.08}          \\
  & $+$ ID ETC               & \underline{45.83}          & \multicolumn{2}{c}{27.08}          \\
  & $+$ OOD ETC (Human)             & 37.50         & \multicolumn{2}{c}{\textbf{50.00}} \\
  & $+$ OOD ETC (GPT-Image-2)       & 38.89          & \multicolumn{2}{c}{\underline{37.50}}          \\
\bottomrule
\end{tabular}
\end{table}

We then validate whether this effect extends to real-robot deployment. To isolate the contribution of downstream ETC, all variants start from policies trained with both Distribution Bridging and Objective Bridging and differ only in their Stage3 supervision (Table\ref{tab:realrobot_compositional_induction}). On the coffee-bag task, OOD ETC improves OOD performance from $41.67$ under action-only adaptation to $80.56$, and further outperforms in-domain ETC ($69.44$). On the flower task, in-domain ETC yields no improvement over action-only adaptation ($27.08$ for both), whereas OOD ETC raises OOD performance to $50.00$. These results indicate that ETC must cover the unseen deployment conditions, rather than merely augmenting in-domain supervision. Consistently, the attention heatmaps in Figure~\ref{fig:retentive}b show that the policy without OOD ETC attends to distractors, such as the wrong bag or flower, whereas OOD ETC redirects attention toward the requested unseen target.

Together, the simulation and real-robot results demonstrate that targeted OOD ETC composes unseen visual–language conditions with previously acquired manipulation behaviors, enabling generalization without corresponding OOD action demonstrations.
% We next examine whether compositional induction transfers from simulation to real-robot deployment. To isolate the effect of downstream ETC, all variants start from policies trained with both Distribution Bridging and Objective Bridging, and differ only in their Stage~3 supervision (Table~\ref{tab:realrobot_compositional_induction}). On both tasks, OOD ETC achieves the best OOD score. On the coffee-bag task, OOD ETC substantially improves OOD performance from $41.67$ under action-only to $80.56$, and notably surpasses in-domain ETC ($69.44$), indicating that ETC must target unseen conditions rather than merely augment in-domain supervision. The flower task makes this even more striking: in-domain ETC yields no OOD improvement over action-only ($27.08$ for both), whereas OOD ETC nearly doubles OOD performance to $50.00$. The attention heatmaps in Figure~\ref{fig:retentive}b also show that the model without OOD ETC grounds to distractors like the wrong bag or the wrong flower, whereas OOD ETC redirects attention to the requested unseen target. Together, these results show that ETC targeted at unseen deployment conditions can connect new visual-language compositions with previously acquired manipulation behaviors, without requiring corresponding OOD action demonstrations.
% \vspace{-0.3em}

\subsubsection{Cost-Efficient Construction of OOD ETC.}
Effective ETC for novel conditions need not rely on costly manual annotation or additional robot demonstrations. In the coffee-bag task, Gemini-generated annotations on real-world images outperform human annotations, improving the score from $80.56$ to $\textbf{91.67}$. In the flower task, photorealistic target-condition images generated by GPT-Image-2 enable ETC construction without collecting new visual observations, improving the score from $27.08$ to $37.50$, although still below the real-image, human-annotated setting of $50.00$. Together, automated annotation and generative visual construction provide a scalable route to compositional induction without new action demonstrations.

% \vspace{-0.3em}
\section{Conclusion}

We formulate VLM-to-VLA adaptation as a dual mismatch in input distribution and training objective, and propose an ETC-guided three-stage pathway to bridge these gaps. Our results show that ETC provides a stronger VLA initialization, remains more effective than general multimodal supervision during action learning, and supports reliable adaptation on real-robot deployment tasks. Experiments in simulation and on real robots demonstrate that this pathway improves generalizable policy learning. Beyond this staged adaptation, low-cost OOD ETC enables \emph{compositional induction}, allowing policies to generalize to unseen visual-language compositions without collecting corresponding OOD action demonstrations. 

% \vspace{-0.3em}
\section{Limitations}
The three-stage pipeline is more compute-intensive than direct VLA fine-tuning, since each stage adds training over the VLM backbone. Our real-robot validation is also restricted to a single WidowX arm, leaving cross-embodiment transfer an open question.

\clearpage
\bibliographystyle{unsrtnat}
\bibliography{main}

@article{beyer2024paligemma,
  title={Paligemma: A versatile 3b vlm for transfer},
  author={Beyer, Lucas and Steiner, Andreas and Pinto, Andr{\'e} Susano and Kolesnikov, Alexander and Wang, Xiao and Salz, Daniel and Neumann, Maxim and Alabdulmohsin, Ibrahim and Tschannen, Michael and Bugliarello, Emanuele and others},
  journal={arXiv preprint arXiv:2407.07726},
  year={2024}
}

@article{driess2023palm,
  title={Palm-e: An embodied multimodal language model},
  author={Driess, Danny and Xia, Fei and Sajjadi, Mehdi SM and Lynch, Corey and Chowdhery, Aakanksha and Ichter, Brian and Wahid, Ayzaan and Tompson, Jonathan and Vuong, Quan and Yu, Tianhe and others},
  journal={arXiv preprint arXiv:2303.03378},
  year={2023}
}

@inproceedings{zitkovich2023rt,
  title={Rt-2: Vision-language-action models transfer web knowledge to robotic control},
  author={Zitkovich, Brianna and Yu, Tianhe and Xu, Sichun and Xu, Peng and Xiao, Ted and Xia, Fei and Wu, Jialin and Wohlhart, Paul and Welker, Stefan and Wahid, Ayzaan and others},
  booktitle={Conference on Robot Learning},
  pages={2165--2183},
  year={2023},
  organization={PMLR}
}

@article{pi0,
  title={{$\pi_0$}: A Vision-Language-Action Flow Model for General Robot Control},
  author={Black, Kevin and Brown, Noah and Driess, Danny and Esmail, Adnan and Equi, Michael and Finn, Chelsea and Fusai, Niccolo and Groom, Lachy and Hausman, Karol and Ichter, Brian and others},
  journal={arXiv preprint arXiv:2410.24164},
  year={2024}
}

@article{pertsch2025fastefficientactiontokenization,
  title={Fast: Efficient action tokenization for vision-language-action models},
  author={Pertsch, Karl and Stachowicz, Kyle and Ichter, Brian and Driess, Danny and Nair, Suraj and Vuong, Quan and Mees, Oier and Finn, Chelsea and Levine, Sergey},
  journal={arXiv preprint arXiv:2501.09747},
  year={2025}
}

@misc{pi05,
      title={$\pi_{0.5}$: a Vision-Language-Action Model with Open-World Generalization}, 
      author={Physical Intelligence and Kevin Black and Noah Brown and James Darpinian and Karan Dhabalia and Danny Driess and Adnan Esmail and Michael Equi and Chelsea Finn and Niccolo Fusai and Manuel Y. Galliker and Dibya Ghosh and Lachy Groom and Karol Hausman and Brian Ichter and Szymon Jakubczak and Tim Jones and Liyiming Ke and Devin LeBlanc and Sergey Levine and Adrian Li-Bell and Mohith Mothukuri and Suraj Nair and Karl Pertsch and Allen Z. Ren and Lucy Xiaoyang Shi and Laura Smith and Jost Tobias Springenberg and Kyle Stachowicz and James Tanner and Quan Vuong and Homer Walke and Anna Walling and Haohuan Wang and Lili Yu and Ury Zhilinsky},
      year={2025},
      eprint={2504.16054},
      archivePrefix={arXiv},
      primaryClass={cs.LG},
      url={https://arxiv.org/abs/2504.16054}, 
}

@article{gr00t_n1,
  title={Gr00t n1: An open foundation model for generalist humanoid robots},
  author={Bjorck, Johan and Casta{\~n}eda, Fernando and Cherniadev, Nikita and Da, Xingye and Ding, Runyu and Fan, Linxi and Fang, Yu and Fox, Dieter and Hu, Fengyuan and Huang, Spencer and others},
  journal={arXiv preprint arXiv:2503.14734},
  year={2025}
}

@inproceedings{embspatial,
  title={Embspatial-bench: Benchmarking spatial understanding for embodied tasks with large vision-language models},
  author={Du, Mengfei and Wu, Binhao and Li, Zejun and Huang, Xuan-Jing and Wei, Zhongyu},
  booktitle={Proceedings of the 62nd Annual Meeting of the Association for Computational Linguistics (Volume 2: Short Papers)},
  pages={346--355},
  year={2024}
}

@article{robopoint,
  title={Robopoint: A vision-language model for spatial affordance prediction for robotics},
  author={Yuan, Wentao and Duan, Jiafei and Blukis, Valts and Pumacay, Wilbert and Krishna, Ranjay and Murali, Adithyavairavan and Mousavian, Arsalan and Fox, Dieter},
  journal={arXiv preprint arXiv:2406.10721},
  year={2024}
}

@article{roborefer,
  title={Roborefer: Towards spatial referring with reasoning in vision-language models for robotics},
  author={Zhou, Enshen and An, Jingkun and Chi, Cheng and Han, Yi and Rong, Shanyu and Zhang, Chi and Wang, Pengwei and Wang, Zhongyuan and Huang, Tiejun and Sheng, Lu and others},
  journal={Advances in Neural Information Processing Systems},
  volume={38},
  pages={28404--28481},
  year={2026}
}

@inproceedings{robovqa,
  title={Robovqa: Multimodal long-horizon reasoning for robotics},
  author={Sermanet, Pierre and Ding, Tianli and Zhao, Jeffrey and Xia, Fei and Dwibedi, Debidatta and Gopalakrishnan, Keerthana and Chan, Christine and Dulac-Arnold, Gabriel and Maddineni, Sharath and Joshi, Nikhil J and others},
  booktitle={2024 IEEE International Conference on Robotics and Automation (ICRA)},
  pages={645--652},
  year={2024},
  organization={IEEE}
}

@inproceedings{robobrain,
  title={Robobrain: A unified brain model for robotic manipulation from abstract to concrete},
  author={Ji, Yuheng and Tan, Huajie and Shi, Jiayu and Hao, Xiaoshuai and Zhang, Yuan and Zhang, Hengyuan and Wang, Pengwei and Zhao, Mengdi and Mu, Yao and An, Pengju and others},
  booktitle={Proceedings of the IEEE/CVF Conference on Computer Vision and Pattern Recognition},
  pages={1724--1734},
  year={2025}
}

@article{robobrain2,
  title={Robobrain 2.0 technical report},
  author={Team, BAAI RoboBrain and Cao, Mingyu and Tan, Huajie and Ji, Yuheng and Chen, Xiansheng and Lin, Minglan and Li, Zhiyu and Cao, Zhou and Wang, Pengwei and Zhou, Enshen and others},
  journal={arXiv preprint arXiv:2507.02029},
  year={2025}
}

@article{vlaos,
  title={Vla-os: Structuring and dissecting planning representations and paradigms in vision-language-action models},
  author={Gao, Chongkai and Liu, Zixuan and Chi, Zhenghao and Huang, Junshan and Fei, Xin and Hou, Yiwen and Zhang, Yuxuan and Lin, Yudi and Fang, Zhirui and Shao, Lin},
  journal={Advances in Neural Information Processing Systems},
  volume={38},
  pages={136705--136736},
  year={2026}
}

@article{molmoact,
  title={Molmoact: Action reasoning models that can reason in space},
  author={Lee, Jason and Duan, Jiafei and Fang, Haoquan and Deng, Yuquan and Liu, Shuo and Li, Boyang and Fang, Bohan and Zhang, Jieyu and Wang, Yi Ru and Lee, Sangho and others},
  journal={arXiv preprint arXiv:2508.07917},
  year={2025}
}

@Article{robo2vlm,
  Title                    = {{Robo2VLM}: Visual Question Answering from Large-Scale In-the-Wild Robot Manipulation Datasets},
  Author                   = {Chen, Kaiyuan and Xie, Shuangyu and Ma, Zehan and Sanketi, Pannag R. and Goldberg, Ken},
  Journal                  = {arXiv preprint arXiv:2505.15517},
  Year                     = {2025}
}

@misc{qu2026eo1openunifiedembodied,
      title={EO-1: An Open Unified Embodied Foundation Model for General Robot Control}, 
      author={Delin Qu and Haoming Song and Qizhi Chen and Zhaoqing Chen and Xianqiang Gao and Dong Wang and Xinyi Ye and Qi Lv and Modi Shi and Guanghui Ren and Cheng Ruan and Maoqing Yao and Haoran Yang and Jiacheng Bao and Bin Zhao and Xuelong Li},
      year={2026},
      eprint={2508.21112},
      archivePrefix={arXiv},
      primaryClass={cs.RO},
      url={https://arxiv.org/abs/2508.21112}, 
}

@article{lin2026systematic,
  title={A Systematic Study of Data Modalities and Strategies for Co-training Large Behavior Models for Robot Manipulation},
  author={Lin, Fanqi and Arora, Kushal and Mercat, Jean and Nishimura, Haruki and Shah, Paarth and Xu, Chen and Zhang, Mengchao and Zolotas, Mark and Angeles, Maya and Pfannenstiehl, Owen and others},
  journal={arXiv preprint arXiv:2602.01067},
  year={2026}
}

@article{embodiedmidtrain,
  title={EmbodiedMidtrain: Bridging the Gap between Vision-Language Models and Vision-Language-Action Models via Mid-training},
  author={Du, Yiyang and Guo, Zhanqiu and Ye, Xin and Ren, Liu and Xiong, Chenyan},
  journal={arXiv preprint arXiv:2604.20012},
  year={2026}
}

@article{vlm2vla,
  title={Actions as language: Fine-tuning vlms into vlas without catastrophic forgetting},
  author={Hancock, Asher J and Wu, Xindi and Zha, Lihan and Russakovsky, Olga and Majumdar, Anirudha},
  journal={arXiv preprint arXiv:2509.22195},
  year={2025}
}

@article{zhang2026vlm4vlarevisitingvisionlanguagemodelsvisionlanguageaction,
  title={VLM4VLA: Revisiting Vision-Language-Models in Vision-Language-Action Models},
  author={Zhang, Jianke and Chen, Xiaoyu and Wang, Qiuyue and Li, Mingsheng and Guo, Yanjiang and Hu, Yucheng and Zhang, Jiajun and Bai, Shuai and Lin, Junyang and Chen, Jianyu},
  journal={arXiv preprint arXiv:2601.03309},
  year={2026}
}

@inproceedings{chatvla,
  title={Chatvla: Unified multimodal understanding and robot control with vision-language-action model},
  author={Zhou, Zhongyi and Zhu, Yichen and Zhu, Minjie and Wen, Junjie and Liu, Ning and Xu, Zhiyuan and Meng, Weibin and Peng, Yaxin and Shen, Chaomin and Feng, Feifei and others},
  booktitle={Proceedings of the 2025 Conference on Empirical Methods in Natural Language Processing},
  pages={5377--5395},
  year={2025}
}

@article{knowledge_insulating_vla,
  title={Knowledge insulating vision-language-action models: Train fast, run fast, generalize better},
  author={Driess, Danny and Springenberg, Jost and Ichter, Brian and Yu, Lili and Li-Bell, Adrian and Pertsch, Karl and Ren, Allen and Walke, Homer and Vuong, Quan and Shi, Lucy Xiaoyang and others},
  journal={Advances in Neural Information Processing Systems},
  volume={38},
  pages={102867--102888},
  year={2026}
}

@inproceedings{vlabench,
  title={Vlabench: A large-scale benchmark for language-conditioned robotics manipulation with long-horizon reasoning tasks},
  author={Zhang, Shiduo and Xu, Zhe and Liu, Peiju and Yu, Xiaopeng and Li, Yuan and Gao, Qinghui and Fei, Zhaoye and Yin, Zhangyue and Wu, Zuxuan and Jiang, Yu-Gang and others},
  booktitle={Proceedings of the IEEE/CVF International Conference on Computer Vision},
  pages={11142--11152},
  year={2025}
}

@article{libero,
  title={Libero: Benchmarking knowledge transfer for lifelong robot learning},
  author={Liu, Bo and Zhu, Yifeng and Gao, Chongkai and Feng, Yihao and Liu, Qiang and Zhu, Yuke and Stone, Peter},
  journal={Advances in Neural Information Processing Systems},
  volume={36},
  pages={44776--44791},
  year={2023}
}

@article{simpler,
  title={Evaluating real-world robot manipulation policies in simulation},
  author={Li, Xuanlin and Hsu, Kyle and Gu, Jiayuan and Pertsch, Karl and Mees, Oier and Walke, Homer Rich and Fu, Chuyuan and Lunawat, Ishikaa and Sieh, Isabel and Kirmani, Sean and others},
  journal={arXiv preprint arXiv:2405.05941},
  year={2024}
}

@article{gemini_robotics,
  title={Gemini robotics: Bringing ai into the physical world},
  author={Team, Gemini Robotics and Abeyruwan, Saminda and Ainslie, Joshua and Alayrac, Jean-Baptiste and Arenas, Montserrat Gonzalez and Armstrong, Travis and Balakrishna, Ashwin and Baruch, Robert and Bauza, Maria and Blokzijl, Michiel and others},
  journal={arXiv preprint arXiv:2503.20020},
  year={2025}
}

@inproceedings{a_okvqa,
  title={A-okvqa: A benchmark for visual question answering using world knowledge},
  author={Schwenk, Dustin and Khandelwal, Apoorv and Clark, Christopher and Marino, Kenneth and Mottaghi, Roozbeh},
  booktitle={European conference on computer vision},
  pages={146--162},
  year={2022},
  organization={Springer}
}

@misc{latex_ocr,
  author       = {Blecher, Lukas},
  title        = {{LaTeX-OCR}: pix2tex -- Using a {ViT} to Convert Images of Equations into {LaTeX} Code},
  howpublished = {\url{https://github.com/lukas-blecher/LaTeX-OCR}},
  year         = {2022},
  note         = {Software repository, accessed 2026-05-28}
}

@article{xiaomi_robotics_0,
  title={Xiaomi-robotics-0: An open-sourced vision-language-action model with real-time execution},
  author={Cai, Rui and Guo, Jun and He, Xinze and Jin, Piaopiao and Li, Jie and Lin, Bingxuan and Liu, Futeng and Liu, Wei and Ma, Fei and Ma, Kun and others},
  journal={arXiv preprint arXiv:2602.12684},
  year={2026}
}

@article{galaxea_g0,
  title={Galaxea open-world dataset and g0 dual-system vla model},
  author={Jiang, Tao and Yuan, Tianyuan and Liu, Yicheng and Lu, Chenhao and Cui, Jianning and Liu, Xiao and Cheng, Shuiqi and Gao, Jiyang and Xu, Huazhe and Zhao, Hang},
  journal={arXiv preprint arXiv:2509.00576},
  year={2025}
}

@article{gr3,
  title={Gr-3 technical report},
  author={Cheang, Chilam and Chen, Sijin and Cui, Zhongren and Hu, Yingdong and Huang, Liqun and Kong, Tao and Li, Hang and Li, Yifeng and Liu, Yuxiao and Ma, Xiao and others},
  journal={arXiv preprint arXiv:2507.15493},
  year={2025}
}

@article{internvla_a1,
  title={InternVLA-A1: Unifying Understanding, Generation and Action for Robotic Manipulation},
  author={Cai, Junhao and Cai, Zetao and Cao, Jiafei and Chen, Yilun and He, Zeyu and Jiang, Lei and Li, Hang and Li, Hengjie and Li, Yang and Liu, Yufei and others},
  journal={arXiv preprint arXiv:2601.02456},
  year={2026}
}

@article{mimo_embodied,
  title={Mimo-embodied: X-embodied foundation model technical report},
  author={Hao, Xiaoshuai and Zhou, Lei and Huang, Zhijian and Hou, Zhiwen and Tang, Yingbo and Zhang, Lingfeng and Li, Guang and Lu, Zheng and Ren, Shuhuai and Meng, Xianhui and others},
  journal={arXiv preprint arXiv:2511.16518},
  year={2025}
}

@article{liu2025faster,
  title={FASTer: Toward Efficient Autoregressive Vision Language Action Modeling via Neural Action Tokenization},
  author={Liu, Yicheng and Zhang, Shiduo and Dong, Zibin and Ye, Baijun and Yuan, Tianyuan and Yu, Xiaopeng and Yin, Linqi and Lu, Chenhao and Shi, Junhao and Yu, Luca Jiang-Tao and others},
  journal={arXiv preprint arXiv:2512.04952},
  year={2025}
}

@article{zhai2025ignitingvlmsembodiedspace,
  title={Igniting vlms toward the embodied space},
  author={Zhai, Andy and Liu, Brae and Fang, Bruno and Cai, Chalse and Ma, Ellie and Yin, Ethan and Wang, Hao and Zhou, Hugo and Wang, James and Shi, Lights and others},
  journal={arXiv preprint arXiv:2509.11766},
  year={2025}
}

@article{wu2026pragmatic,
  title={A Pragmatic VLA Foundation Model},
  author={Wu, Wei and Lu, Fan and Wang, Yunnan and Yang, Shuai and Liu, Shi and Wang, Fangjing and Zhu, Qian and Sun, He and Wang, Yong and Ma, Shuailei and others},
  journal={arXiv preprint arXiv:2601.18692},
  year={2026}
}

@misc{xie2026unifyrobotactionscamera,
      title={Unify Robot Actions in Camera Frame}, 
      author={Sicheng Xie and Lingchen Meng and Zijie Diao and Haidong Cao and Zhiying Du and Shuyuan Tu and Jiaqi Leng and Qiuyue Wang and Mingsheng Li and Shuai Bai and Zuxuan Wu and Yu-Gang Jiang},
      year={2026},
      eprint={2511.17001},
      archivePrefix={arXiv},
      primaryClass={cs.RO},
      url={https://arxiv.org/abs/2511.17001}, 
}

@inproceedings{walke2023bridgedata,
  title={Bridgedata v2: A dataset for robot learning at scale},
  author={Walke, Homer Rich and Black, Kevin and Zhao, Tony Z and Vuong, Quan and Zheng, Chongyi and Hansen-Estruch, Philippe and He, Andre Wang and Myers, Vivek and Kim, Moo Jin and Du, Max and others},
  booktitle={Conference on Robot Learning},
  pages={1723--1736},
  year={2023},
  organization={PMLR}
}

@inproceedings{lin2014microsoft,
  title={Microsoft coco: Common objects in context},
  author={Lin, Tsung-Yi and Maire, Michael and Belongie, Serge and Hays, James and Perona, Pietro and Ramanan, Deva and Doll{\'a}r, Piotr and Zitnick, C Lawrence},
  booktitle={European conference on computer vision},
  pages={740--755},
  year={2014},
  organization={Springer}
}

\clearpage
\appendix
\section{Implementation Details}
\label{sec:implementation_details}

\subsection{Co-training Strategy}
\label{subsec:cotrain_strategy}
During both Objective Bridging and Retentive Adaptation, we co-train on ETC and action data. Each optimization step combines one action batch with one ETC batch: we compute the action loss and ETC loss in separate forward/backward passes, accumulate or explicitly sum their gradients, and then perform a single optimizer update on the combined gradients. This decouples the two objective evaluations while keeping a joint update, reducing interference between action learning and ETC supervision throughout co-training.

\subsection{Hyperparameters}
\label{subsec:hyperparams}

\noindent\textbf{Stage~1 (Distribution Bridging).}
All backbones are fine-tuned on ETC supervision using the ms-swift framework with full-parameter fine-tuning (ViT included, \texttt{freeze\_vit=false}), DeepSpeed ZeRO-3, and BF16 precision.
Key hyperparameters are given in Table~\ref{tab:stage1_hparams}.

\begin{table}[h]
  \centering
  \caption{Stage~1 ETC pretraining hyperparameters (shared across all source datasets).}
  \label{tab:stage1_hparams}
  \small
  \setlength{\tabcolsep}{5pt}
  \begin{tabular}{lcc}
    \toprule
    Hyperparameter & PaliGemma-3B & Qwen2.5-VL-3B \\
    \midrule
    Max steps & 20{,}000 & 20{,}000 \\
    Learning rate & $1\!\times\!10^{-5}$ & $1\!\times\!10^{-5}$ \\
    Warmup ratio & 0.05 & 0.05 \\
    LR schedule & cosine & cosine \\
    Per-device batch size & 2 & 2 \\
    Gradient accumulation & 1 & 1 \\
    Max sequence length & 8{,}192 & 8{,}192 \\
    Sequence packing & $\checkmark$ & $\checkmark$ \\
    Precision & BF16 & BF16 \\
    \bottomrule
  \end{tabular}
\end{table}

\noindent\textbf{Stage~2 (Objective Bridging).}
Bridge and LIBERO experiments use the AR-VLA-cotraining framework with a PiFAST backbone;
VLABench experiments use the Pi0-FAST framework.
Both share the cosine-decay optimizer schedule; differences are listed in Table~\ref{tab:stage2_hparams}.

\begin{table}[h]
  \centering
  \caption{Stage~2 Objective Bridging hyperparameters.}
  \label{tab:stage2_hparams}
  \small
  \setlength{\tabcolsep}{4pt}
  \begin{tabular}{lccc}
    \toprule
    Hyperparameter & Bridge & LIBERO & VLABench \\
    \midrule
    Architecture & PiFAST & PiFAST & Pi0-FAST \\
    Stage~1 init & ShareRobot & VLA-OS & VLABench VQA \\
    Camera views & 1 (primary) & 2 (primary + wrist) & 3 (primary + wrist + second image) \\
    Max steps & 100{,}000 & 30{,}000 & 100{,}000 \\
    Learning rate & $2.5\!\times\!10^{-5}$ & $2.5\!\times\!10^{-5}$ & $2.5\!\times\!10^{-5}$ \\
    Warmup steps & 1{,}000 & 1{,}000 & 1{,}000 \\
    LR schedule & cosine & cosine & cosine \\
    Global batch size & 128 & 128 & 32 \\
    Optimizer & AdamW (8-bit) & AdamW (8-bit) & AdamW \\
    Action horizon & 10 & 10 & 10 \\
    Action dim & 7 (EEF delta) & 7 (EEF delta) & 7 (EEF delta) \\
    ETC co-train ratio & 0.25 & 0.25 & 0.25 \\
    Image augmentation & rot.$\pm$5\textdegree, crop, color jitter & rot.$\pm$5\textdegree, crop, color jitter & — \\
    Action chunk encoding & rel. to current & rel. to current & step-wise delta \\
    \bottomrule
  \end{tabular}
\end{table}

\subsection{Data Mixing Strategy}
\label{subsec:data_mixing}

\noindent\textbf{Stage~1.} For each ETC source, we uniformly sample its sub-categories to maintain a balanced distribution across sub-types.

\noindent\textbf{Objective Bridging.} We use the joint-update co-training schedule described in Section~\ref{subsec:cotrain_strategy}. The ETC batch size is set to 25\% of the action batch size (\texttt{cotrain\_ratio}$=0.25$), and ETC sub-categories are sampled uniformly. In the $+$MM ablations on Bridge and LIBERO, the ETC stream is replaced by generic MM data (A-OKVQA, COCO, and LaTeX-OCR) at the same 25\% ratio, with the three MM datasets mixed uniformly.

\subsection{Hardware and Compute}

\noindent\textbf{Training.}
Stage~1 ETC pretraining is conducted on 4 NVIDIA H200 141G GPUs.
Objective Bridging is conducted on 8 NVIDIA H200 141G GPUs.
All training runs use BF16 mixed precision.

\noindent\textbf{Evaluation.}
All simulation evaluations are conducted on NVIDIA H100 80G GPUs (CUDA 12.8).
All models use FP32 precision during inference.

\section{Additional Backbone Results}
\label{sec:backbone_ablation}

The main experiments focus on PaliGemma backbones.  Here we report additional
backbone ablations using (i) \textbf{SigLIP+Gemma}, a PiFAST model initialized
from a raw SigLIP-400M vision encoder and Gemma-2B LLM without joint VLM
pretraining, and (ii) \textbf{Qwen2.5-VL-3B}, evaluated with and without
Stage~1 ETC pretraining where runs are available.

\begin{table}[h]
  \centering
  \caption{SigLIP+Gemma backbone ablation under action-only and co-training
           conditions.  LP-train and LB-train denote LIBERO-Plus evaluation for
           policies trained on LIBERO-Plus and LIBERO, respectively.}
  \label{tab:siglip_ablation}
  \small
  \setlength{\tabcolsep}{4pt}
  \begin{tabular}{lcccc}
    \toprule
    Cotrain data & Bridge & LIBERO & LP-train & LB-train \\
    \midrule
    Action only  & \textbf{0.385} & \textbf{91.5\%} & 0.324 & 0.406 \\
    $+$ MM       & 0.021 & 84.6\% & \textbf{0.397} & 0.379 \\
    $+$ ETC      & 0.177 & 89.4\% & 0.367 & 0.382 \\
    $+$ MM+ETC   & 0.188 & 85.9\% & — & \textbf{0.437} \\
    \bottomrule
  \end{tabular}
\end{table}

\begin{table}[h]
  \centering
  \caption{Qwen2.5-VL-3B backbone results on LIBERO.}
  \label{tab:qwen_ablation}
  \small
  \setlength{\tabcolsep}{5pt}
  \begin{tabular}{llc}
    \toprule
    Init & Cotrain data & LIBERO \\
    \midrule
    \multirow{4}{*}{Qwen2.5-VL-3B (w/o ETC)}
      & Action only & 74.3\% \\
      & $+$ MM & 73.2\% \\
      & $+$ ETC & \textbf{78.2}\% \\
      & $+$ MM+ETC & 70.0\% \\
    \midrule
    \multirow{4}{*}{Qwen2.5-VL-3B (w/ ETC)}
      & Action only & 66.6\% \\
      & $+$ MM & 72.0\% \\
      & $+$ ETC & \textbf{73.2}\% \\
      & $+$ MM+ETC & 72.6\% \\
    \bottomrule
  \end{tabular}
\end{table}

\section{Data}
\label{sec:data}

\subsection{ETC Data Curation Pipeline}
\label{sec:vqa_construction_pipeline}

Our VQA pipeline grounds questions in real-robot recorded trajectories, inspired
by the Robo2VLM paradigm of leveraging synchronized proprioceptive and visual
modalities.  Ground-truth answers are produced from structured trajectory
signals or Gemini-2.5-Pro generation and are manually verified.  For most
question types, a chain of thought is generated and merged with the answer
during training.  The pipeline covers the following five reasoning categories:

Trajectory Prediction. Given the initial frame and a language instruction, the model predicts five equidistant waypoints along the recorded end-effector trajectory to abstract the full manipulation motion, evaluating the ability to anticipate complete motion from a single image.

Spatial Understanding. View Correspondence requires the model to match a BBox-highlighted object in one viewpoint to its numeric ID in another, demanding cross-view geometric reasoning without depth or calibration data. Relative Direction asks the model to determine the 3D spatial relation between the target object and the end-effector, choosing from {left, right, forward, backward, upper, lower}. Directions are defined in a gripper-centric frame: left/right along the gripper’s local x-axis (perpendicular to the opening plane), forward/backward along the local y-axis (approach direction), and upper/lower along the world z-axis (vertical displacement).

Affordance Reasoning. Affordance Validation presents a highlighted grasp region and asks for a Yes/No judgment on grasp stability. Affordance Localization requires regressing the optimal grasp point for a specified target object.

Goal-Conditioned Reasoning. Target Object Identification selects the numeric ID of the task-specified object from an annotated initial frame. Target Object Localization requires regressing a bounding box for that object. Target Object Description generates a natural-language description of the object’s manipulation-relevant (and in some cases, semantic) attributes, testing open-domain visual recognition.

Task Planning. Action Understanding classifies the current manipulation phase from the onset frame and task instruction. Predict Action forecasts the next action from an end-of-phase frame. Sub-task Sequencing generates a complete natural-language action plan from the initial frame and full instruction, evaluating long-horizon procedural reasoning.

Table~\ref{tab:vqa_construction_examples} summarizes the question prototypes
used to construct VQA examples for each ETC capability, together with the source
of the ground-truth answer.

\begin{table*}[t]
  \centering
  \caption{VQA construction examples grouped by capability.}
  \label{tab:vqa_construction_examples}
  \scriptsize
  \setlength{\tabcolsep}{4pt}
  \renewcommand{\arraystretch}{1.12}

    \begin{tabularx}{\textwidth}{
      @{}
      >{\ttfamily\raggedright\arraybackslash}l
      >{\raggedright\arraybackslash}X
      >{\raggedright\arraybackslash}p{0.20\textwidth}
      @{}
    }
    \toprule
    {\normalfont\bfseries Capability} & {\bfseries Question Prototype} & {\bfseries Ground-truth Source} \\
    \midrule
    \groupheader{Trajectory Prediction}

    \makecell[tl]{\texttt{subtask\_trajectory\_prediction}}
      & Predict the coordinates of five key points of the robot’s gripper trajectory. Output as [(x1,y1), (x2,y2), (x3,y3), (x4,y4), (x5,y5)]. The robot is asked to \{instruction\}.
      & Computed from real robot data / generated by Gemini \\

    \groupheader{Spatial Understanding}

    \makecell[tl]{\texttt{view\_correspondence}}
      & The first image includes the bounding box of a specific object. Which bounding box in the second image corresponds to the same object, based on attributes relevant for manipulation? Answer the correct ID.
      & Generated by Gemini \\

    \makecell[tl]{\texttt{relative\_direction}}
      & In the image, which direction is the \{target object\} relative to the robot’s end effector?
      & Computed from real robot data \\

    \groupheader{Affordance}

    \makecell[tl]{\texttt{affordance\_validation}}
      & Given the bounding box, is this region suitable for a stable grasp? Answer yes or no.
      & Generated by Gemini \\

    \makecell[tl]{\texttt{affordance\_localization}}
      & Where is the optimal grasp point for \{target object\}? Provide the pixel coordinates as [x, y].
      & Computed from real robot data \\

    \groupheader{Object Description}

    \makecell[tl]{\texttt{target\_object\_identification}}
      & Which object in the image is the target object? Answer the ID number of its bounding box. The robot is asked to \{instruction\}.
      & Generated by Gemini \\

    \makecell[tl]{\texttt{target\_object\_localization}}
      & In the image, locate the \{target object\} by providing its bounding box. Output the bounding box as [x\_min, y\_min, x\_max, y\_max] in image pixel coordinates.
      & Generated by Gemini \\

    \makecell[tl]{\texttt{target\_object\_description}}
      & Describe the target object of the task. The robot is asked to \{instruction\}. Focus on attributes relevant for manipulation, such as type, shape, color, position, and orientation. Keep your answer concise.
      & Generated by Gemini \\

    \groupheader{Task Planning}

    \makecell[tl]{\texttt{action\_understanding}}
      & The robot is asked to \{instruction\}. Which phase of the grasp action is shown? Choose from [pre grasp, contact, immobilization, detach, post grasp].
      & Computed from real robot data \\

    \makecell[tl]{\texttt{predict\_action}}
      & After the action shown in the image, what will be the robot’s next action for the task \{instruction\}?
      & Generated by Gemini \\

    \makecell[tl]{\texttt{subtasks\_sequencing}}
      & Please describe the sequence of actions the robot will perform to complete the task: \{instruction\}. Keep your answer concise.
      & Generated by Gemini \\

    \bottomrule
  \end{tabularx}
\end{table*}

\subsection{Simulation ETC Data Sources and Statistics}
\label{sec:data_statistics}

For simulation experiments, we construct ETC supervision from three source datasets:
ShareRobot~\citep{robobrain} (Bridge experiments),
VLA-OS~\citep{vlaos} (LIBERO experiments),
and VLABench VQA annotations derived via the pipeline described above
(VLABench experiments; Track~1 in-domain and Track~2 OOD).
Tables~\ref{tab:etc_sharerobot}--\ref{tab:etc_vlabench} report the category and sub-type breakdown for each source.

% ---- ShareRobot ----
\begin{table}[h]
  \centering
  \caption{ShareRobot ETC data statistics (Bridge experiments). Total: 44,862.}
  \label{tab:etc_sharerobot}
  \small
  \setlength{\tabcolsep}{4pt}
  \begin{tabular}{llr}
    \toprule
    Category & Sub-type & Samples \\
    \midrule
    Affordance & Affordance grounding/validation & 6{,}522 \\
    \midrule
    Trajectory & Trajectory prediction & 6{,}870 \\
    \midrule
    \multirow{10}{*}{Planning}
      & Discriminative affordance (+) & 3{,}147 \\
      & Discriminative affordance ($-$) & 3{,}147 \\
      & Future prediction & 3{,}147 \\
      & Generative affordance & 3{,}147 \\
      & Past description & 3{,}147 \\
      & Planning remaining steps & 3{,}147 \\
      & Planning & 3{,}147 \\
      & Planning with context & 3{,}147 \\
      & Success prediction (+) & 3{,}147 \\
      & Success prediction ($-$) & 3{,}147 \\
    \cmidrule{3-3}
      & \textit{Planning subtotal} & \textit{31{,}470} \\
    \midrule
    \multicolumn{2}{r}{\textbf{Total}} & \textbf{44{,}862} \\
    \bottomrule
  \end{tabular}
\end{table}

% ---- VLA-OS ----
\begin{table}[h]
  \centering
  \caption{VLA-OS ETC data statistics (LIBERO experiments). Total: 388,808.}
  \label{tab:etc_vlaos}
  \small
  \setlength{\tabcolsep}{4pt}
  \begin{tabular}{lr}
    \toprule
    Category & Samples \\
    \midrule
    Object BBox grounding & 143{,}668 \\
    Gripper-state reasoning & 83{,}512 \\
    Move-target prediction & 78{,}117 \\
    Sub-task decomposition & 78{,}117 \\
    Task planning & 5{,}394 \\
    \midrule
    \textbf{Total} & \textbf{388{,}808} \\
    \bottomrule
  \end{tabular}
\end{table}

% ---- VLABench ----
\begin{table}[h]
  \centering
  \caption{VLABench ETC data statistics (VLABench experiments).
    T1 = Track~1 in-domain; T2 = Track~2 OOD.}
  \label{tab:etc_vlabench}
  \small
  \setlength{\tabcolsep}{4pt}
  \begin{tabular}{llrr}
    \toprule
    Category & Sub-type & T1 & T2 \\
    \midrule
    \multirow{3}{*}{Affordance}
      & Affordance validation (+) & 2{,}850 & 1{,}281 \\
      & Affordance validation ($-$) & 2{,}851 & 1{,}281 \\
      & Affordance localization & 28{,}433 & 1{,}281 \\
    \cmidrule{3-4}
      & \textit{Subtotal} & \textit{34{,}134} & \textit{3{,}843} \\
    \midrule
    \multirow{3}{*}{Goal Description}
      & Target object description & 1{,}912 & 430 \\
      & Target object identification & 26{,}577 & 1{,}290 \\
      & Target object localization & 28{,}749 & 1{,}286 \\
    \cmidrule{3-4}
      & \textit{Subtotal} & \textit{57{,}238} & \textit{3{,}006} \\
    \midrule
    \multirow{2}{*}{Spatial Understanding}
      & Relative direction & 1{,}913 & 430 \\
      & View correspondence & 30{,}820 & 930 \\
    \cmidrule{3-4}
      & \textit{Subtotal} & \textit{32{,}733} & \textit{1{,}360} \\
    \midrule
    \multirow{2}{*}{Task Planning}
      & Action understanding & 46{,}209 & 2{,}094 \\
      & Sub-task sequencing & 9{,}583 & 698 \\
    \cmidrule{3-4}
      & \textit{Subtotal} & \textit{55{,}792} & \textit{2{,}792} \\
    \midrule
    Trajectory Prediction & Trajectory waypoints & 30{,}674 & 2{,}081 \\
    \midrule
    \multicolumn{2}{r}{\textbf{Total}} & \textbf{210{,}571} & \textbf{13{,}082} \\
    \bottomrule
  \end{tabular}
\end{table}

\subsection{ETC Data Examples}
\begin{figure}
  \centering
  \includegraphics[
    page=1,
    width=\linewidth,
    trim=0cm 1.8cm 0cm 0.8cm,
    clip
  ]{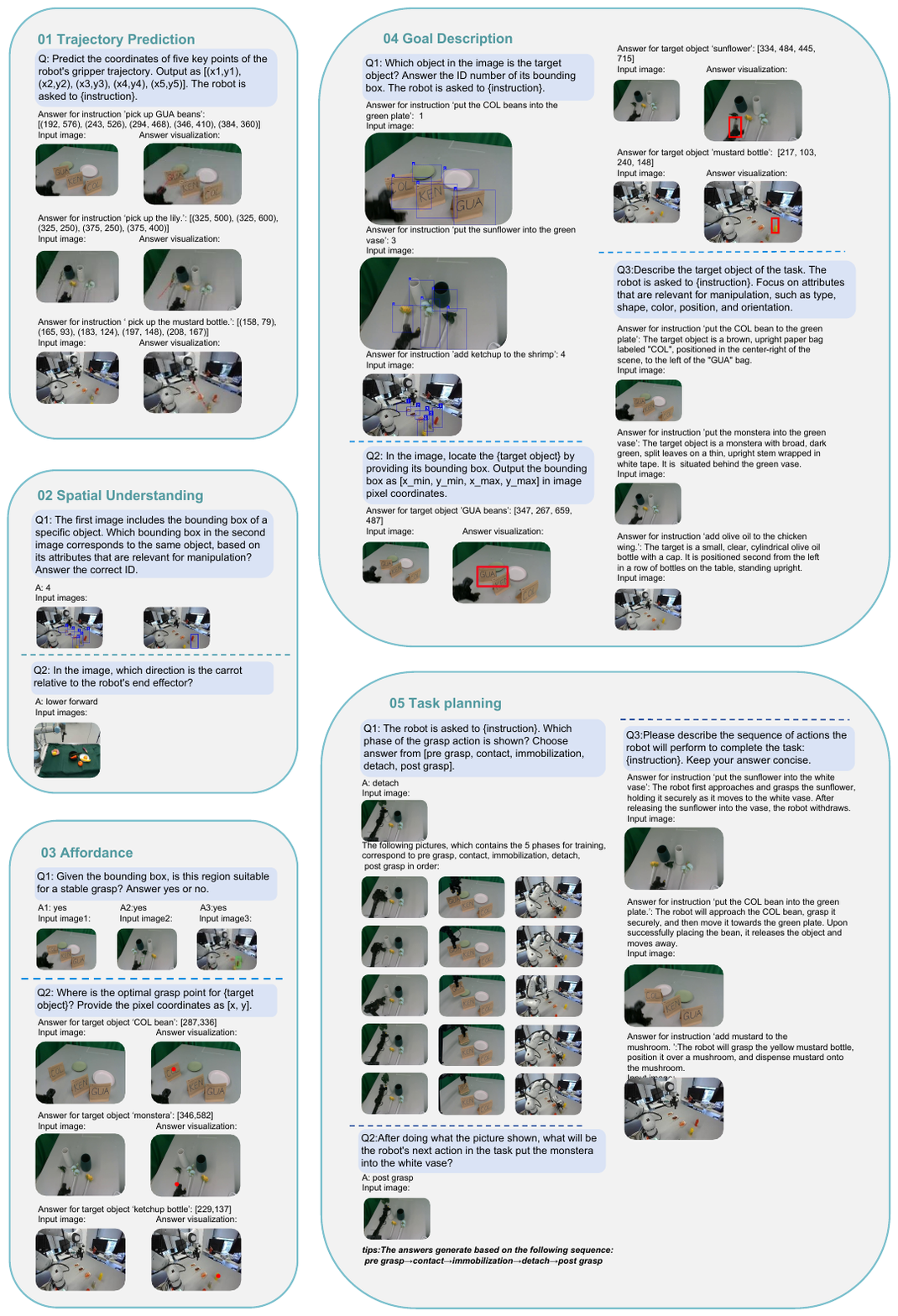}
  \caption{Representative ETC supervision samples constructed from real-robot data. Each row corresponds to a different question type.}
  \label{fig:etc_data_example}
\end{figure}

Figure~\ref{fig:etc_data_example} shows representative ETC supervision samples constructed from real-robot trajectories, covering the question types described in Section~\ref{sec:vqa_construction_pipeline}.
% \begin{figure}
%     \centering
%     \includegraphics[width=0.95\linewidth]{figures/data_example.pdf}
%     \caption{Representative ETC supervision samples constructed from real-robot data. Each row corresponds to a different question type.}
%     \label{fig:etc_data_example}
% \end{figure}

\subsection{MM Data for Ablation}
\label{sec:mm_data}
For the \texttt{$+$ MM} and \texttt{$+$ MM+ETC} ablations in Table~\ref{tab:vla_sr_cotrain},
we use A-OKVQA~\citep{a_okvqa} ($24{,}903$ samples),
COCO object detection~\citep{lin2014microsoft} ($94{,}208$ samples), and
LaTeX-OCR~\citep{latex_ocr} ($192{,}446$ samples)
as generic multimodal supervision, totaling $311{,}557$ samples.

\section{Understanding Evaluation Details}
\label{sec:emm_bench_details}

This section describes how the benchmark scores reported in the main paper are computed, and provides per-category breakdowns for all experimental conditions.
The evaluation covers two groups of benchmarks: \emph{embodied multimodal benchmarks} (ETC), including ShareRobot, VLA-OS, VLABenchVQA, and Robo2VLM---all constructed from the same ETC data sources described in Section~\ref{sec:data_statistics}---and \emph{general multimodal benchmarks} (MM), including LaTeX-OCR, COCO, and A-OKVQA.
Table~\ref{tab:prevlm_all} reports pre-action VLM scores across the four ETC benchmarks;
Table~\ref{tab:midtrain_sharerobot_details} details Objective Bridging co-training results per benchmark.

\paragraph{Evaluation protocol.}
All benchmarks are evaluated in an offline prediction-then-scoring manner.
Each test example contains one or more images, a natural-language instruction or question, and a reference answer.
The model response is decoded deterministically and saved before metric computation.
The scorer assigns each example a continuous score and a binary correctness label; the reported accuracy is the percentage of examples whose binary label is true.
Each dataset is decomposed into task-specific sub-scores so that localization, trajectory prediction, planning, recognition, and open-text generation are reported separately.

\paragraph{Metric convention.}
Unless otherwise noted, reported scores are percentages.  Benchmark averages are
unweighted means over the visible sub-scores in each table.  We keep the paper
tables keyed by semantic experimental conditions rather than internal run IDs:
\emph{Init} specifies the VLM initialization, \emph{Training mix} specifies the
VL co-training source, and \emph{Action data} specifies the robot dataset used
for VLA action training.

%------------------------------------------------------------------
\subsection{Scoring Rules}
\label{sec:scoring_rules}

Several scoring rules are shared across datasets.

\subsubsection{Text Similarity}
For open-ended short answers, planning outputs, and natural-language descriptions, we use token-level F1 unless otherwise specified.
Prediction and reference are lowercased and tokenized; let the multiset overlap be $m$.
Precision $= m / |\text{pred}|$, recall $= m / |\text{ref}|$, and the text score is their harmonic mean.
Empty prediction and empty reference receive score 1; if only one side is empty, the score is 0.

For LaTeX-OCR, we use character-level sequence similarity instead, because correctness is often determined by local character differences (braces, subscripts, symbols).

For spatial-relation questions in VLABenchVQA, we extract a set of relation labels (\texttt{left}, \texttt{right}, \texttt{forward}, \texttt{backward}, \texttt{upper}, \texttt{lower}) from both prediction and reference and compute F1 over the relation sets.
In the reported evaluation the correctness threshold is 1.0 (exact set match).

An example is counted as correct when its best text similarity score reaches the dataset-specific threshold (default 0.8 for short answers; 0.35 for long descriptive VLABenchVQA tasks).

\subsubsection{Normalized Exact Matching}
For yes/no, ID-choice, and VQA-style tasks, answers are normalized (lowercased, punctuation and extra whitespace removed) before comparison.
Yes/no answers are mapped to Boolean values; ID-choice answers are parsed by extracting the first integer.

For Robo2VLM, the evaluator first extracts a final answer from a \texttt{final answer:} field if present, otherwise uses the last non-empty line.
The example is correct if the normalized prediction exactly matches the reference, or if the normalized reference appears as a substring of the prediction.

\subsubsection{Bounding Box IoU}
For BBox localization, predictions and references are parsed as $(x_1, y_1, x_2, y_2)$.
The score is intersection-over-union (IoU); the example is correct if IoU exceeds the task threshold (ShareRobot: 0.5; VLABenchVQA: 0.6; VLA-OS: 0.7).

\subsubsection{Point Distance}
For point localization, the error is the Euclidean pixel distance $d$.
The continuous score is $\exp(-d/\tau)$; the example is correct if $d \le \tau$.
VLA-OS gripper-position tasks use $\tau = 20$ px; VLABenchVQA affordance-point tasks use $\tau = 50$ px.

\subsubsection{Trajectory RMSE}
Predicted and reference trajectories are parsed as sequences of 2D pixel coordinates and resampled to 50 points along arc length if needed.
The trajectory error is the root-mean-squared pixel error over aligned points:
$\text{RMSE} = \sqrt{\text{mean}_i \| p_i - g_i \|_2^2}$.
The continuous score is $\exp(-\text{RMSE}/\tau)$ with $\tau = 20$ px; the example is correct if RMSE $\le 20$ px.

\subsubsection{COCO Joint Detection F1}
For COCO, we use a dataset-specific metric that jointly requires category-label agreement and BBox overlap.
Class names are normalized to the 80 COCO categories (e.g., ``bike'' $\to$ ``bicycle'').
Predictions and ground-truth objects are matched greedily by descending IoU, with each object used at most once.
A \emph{class-aware} match requires class equality and IoU $\ge 0.6$.
Precision, recall, and F1 are computed from global TP/FP/FN across the dataset.
A \emph{BBox-only} variant ignores class labels and uses IoU alone.

%------------------------------------------------------------------
\subsection{Dataset-Specific Metrics}
\label{sec:dataset_metrics}

\subsubsection{ShareRobot}
ShareRobot contains three test subsets: affordance, trajectory, and planning.
Affordance examples are scored by BBox IoU (threshold 0.5).
Trajectory examples are scored by trajectory RMSE (threshold 20 px).
Planning examples are divided into yes/no (normalized exact match) and open-text questions (token-F1, threshold 0.8).
The overall ShareRobot score reported in our tables is the unweighted mean of the affordance, planning, and trajectory sub-scores.

\subsubsection{VLA-OS}
VLA-OS contains five subsets: \texttt{bbox}, \texttt{gripper\_position}, \texttt{planning}, \texttt{subtask}, and \texttt{move}.
BBox examples use BBox IoU (threshold 0.7).
Gripper-position examples use point distance ($\tau = 20$ px).
Planning, subtask, and move examples use token-F1 (threshold 0.7).

\subsubsection{VLABenchVQA}
VLABenchVQA spans affordance, goal description, spatial understanding, task planning, and trajectory prediction.
Point affordance questions use point distance ($\tau = 50$ px); yes/no questions use normalized exact match; text affordance uses token-F1 (0.8).
BBox tasks use IoU (0.6); ID-choice tasks use parsed-ID exact match; long goal descriptions use token-F1 (0.35).
Spatial-relation examples use relation-set F1 (threshold 1.0).
Task-planning examples use token-F1 (0.8); trajectory examples use RMSE (20 px).

\subsubsection{Robo2VLM}
Robo2VLM contains VQA and reasoning subsets, both evaluated with final-answer extraction followed by normalized exact matching (or token-F1 as a continuous score).

\subsubsection{LaTeX-OCR}
LaTeX-OCR evaluates formula transcription.
The evaluator extracts the final answer (preferring an explicit \texttt{<answer>} block, then \texttt{final answer:}, then the last non-empty line) and computes character-level sequence similarity against the reference.
The example is correct if similarity $\ge 0.8$.

\subsubsection{COCO}
COCO is evaluated as open-set object detection with the joint Detection F1 metric (IoU 0.6) described in Section~\ref{sec:scoring_rules}.
The primary score is the global class-aware F1; the stricter per-image exact-match accuracy is reported as a diagnostic.

\subsubsection{A-OKVQA}
A-OKVQA is evaluated as VQA with multiple acceptable answer forms (direct answers, multiple-choice text, and the correct-option field).
The score is the maximum token-F1 over all candidates; the example is correct if the best score $\ge 0.8$.

%------------------------------------------------------------------
\subsection{Aggregation and Reporting}
\label{sec:aggregation}

For all datasets except COCO, each example contributes one binary correctness label to the overall accuracy and one continuous score to the sub-score average.
For COCO, the primary score is the global class-aware F1 computed from accumulated TP/FP/FN rather than per-image accuracy.

Table~\ref{tab:eval_thresholds} summarizes all thresholds used in the reported evaluation.

\begin{table}[h]
  \centering
  \caption{Evaluation thresholds used in the reported results.}
  \label{tab:eval_thresholds}
  \small
  \setlength{\tabcolsep}{5pt}
  \begin{tabular}{lr}
    \toprule
    Metric & Threshold \\
    \midrule
    ShareRobot BBox IoU          & 0.5 \\
    VLABenchVQA BBox IoU         & 0.6 \\
    VLA-OS BBox IoU              & 0.7 \\
    COCO detection IoU           & 0.6 \\
    Trajectory RMSE              & 20 px \\
    VLA-OS gripper point dist.   & 20 px \\
    VLABenchVQA affordance point & 50 px \\
    Short text token-F1          & 0.8 \\
    VLA-OS text token-F1         & 0.7 \\
    VLABenchVQA long text token-F1 & 0.35 \\
    LaTeX-OCR char.\ similarity  & 0.8 \\
    A-OKVQA token-F1             & 0.8 \\
    Spatial relation F1          & 1.0 \\
    \bottomrule
  \end{tabular}
\end{table}

%------------------------------------------------------------------
\subsection{Benchmark Score Tables}
\label{sec:benchmark_tables}

\paragraph{Pre-action VLM scores.}
Table~\ref{tab:prevlm_all} compares backbone VLMs
of different initializations on each benchmark before any action training.

\begin{table}[h]
  \centering
  \caption{Pre-action VLM scores on the four ETC benchmarks before any action training.}
  \label{tab:prevlm_all}
  \small
  % ---- Row 1: ShareRobot | VLA-OS ----
  \begin{minipage}[t]{0.34\linewidth}
    \centering
    \textbf{ShareRobot} (Bridge-side)\\[3pt]
    \setlength{\tabcolsep}{4pt}
    \begin{tabular}{lrrrr}
      \toprule
      Init & Avg. & Plan \\
      \midrule
      PaliGemma-raw    & 16.33  & 16.33  \\
      PaliGemma-Bridge & 26.65  & 26.65  \\
      \bottomrule
    \end{tabular}
  \end{minipage}
  \hfill
  \begin{minipage}[t]{0.63\linewidth}
    \centering
    \textbf{VLA-OS} (LIBERO-side)\\[3pt]
    \setlength{\tabcolsep}{3pt}
    \begin{tabular}{lrrrrrr}
      \toprule
      Init & Avg. & Move & Plan & Subtask & BBox & Grip \\
      \midrule
      PaliGemma-raw    & 12.21 & 17.74 & 22.82 & 20.46 &  0.00 & 0.00 \\
      PaliGemma-LIBERO & 41.21 & 25.02 & 91.44 & 47.18 & 32.42 & 9.96 \\
      \bottomrule
    \end{tabular}
  \end{minipage}

  \vspace{10pt}

  % ---- Row 2: VLABenchVQA | Robo2VLM ----
  \begin{minipage}[t]{0.60\linewidth}
    \centering
    \textbf{VLABenchVQA}\\[3pt]
    \setlength{\tabcolsep}{3pt}
    \begin{tabular}{lrrrrrr}
      \toprule
      Init & Avg. & Aff. & Traj. & Plan & Goal & Spatial \\
      \midrule
      PaliGemma-raw      &  9.55 & 24.67 &  0.00 &  4.92 &  7.13 & 11.04 \\
      PaliGemma-VLABench & 56.38 & 47.63 & 12.94 & 84.09 & 47.54 & 89.69 \\
      \bottomrule
    \end{tabular}
  \end{minipage}
  \hfill
  \begin{minipage}[t]{0.36\linewidth}
    \centering
    \textbf{Robo2VLM-VQA}\\[3pt]
    \setlength{\tabcolsep}{4pt}
    \begin{tabular}{lrr}
      \toprule
      Init & Avg. & VQA \\
      \midrule
      PaliGemma-raw   & 15.15 & 15.15 \\
      PaliGemma-DROID & 69.29 & 69.29 \\
      \bottomrule
    \end{tabular}
  \end{minipage}
\end{table}

\paragraph{Objective Bridging co-training results.}
ShareRobot (Bridge-side) covers affordance reasoning, high-level planning, and trajectory reasoning.
VLA-OS (LIBERO-side) decomposes scene understanding into move-target prediction, task planning,
subtask decomposition, object BBox grounding, and gripper-state reasoning.
The general MM benchmark measures open-domain VL skill retention; A-OKVQA is evaluated as multiple-choice VQA,
COCO as class-aware object detection, and LaTeX-OCR as expression transcription.
For COCO, the main score is the dataset-level class-aware F1 at IoU~0.6; the diagnostics table additionally
reports box-only F1@0.6 (localization only), mIoU over matched boxes, and image-level exact-match accuracy.
Tables~\ref{tab:midtrain_sharerobot_details}--\ref{tab:coco_detection_diagnostics} report all sub-scores.

\begin{table*}[!t]
  \centering
  \caption{Objective Bridging co-training benchmark details.
    All scores are percentages.}
  \label{tab:midtrain_sharerobot_details}

  % ---- ShareRobot ----
  \textbf{ShareRobot} (Bridge; all variants use Bridge as action data)\\[4pt]
  \small\setlength{\tabcolsep}{4pt}
  \begin{tabular}{llrrrr}
    \toprule
    Init & Training mix & Avg. & Aff. & Plan & Traj. \\
    \midrule
    PaliGemma-ETC & Action only  &  0.00 &  0.00 &  0.00 &  0.00 \\
    PaliGemma-ETC & $+$ MM+ETC   & 38.04 &  5.86 & 15.38 & 92.89 \\
    PaliGemma-ETC & $+$ ETC      & 91.92 & 93.72 & 88.73 & 93.31 \\
    PaliGemma     & Action only  &  0.00 &  0.00 &  0.00 &  0.00 \\
    PaliGemma     & $+$ ETC      & 37.57 &  2.93 & 13.54 & 96.23 \\
    PaliGemma     & $+$ MM+ETC   & 69.29 & 94.56 & 20.00 & 93.31 \\
    \bottomrule
  \end{tabular}

  \vspace{8pt}

  % ---- VLA-OS ----
  \phantomsection\label{tab:midtrain_vlaos_details}%
  \textbf{VLA-OS} (LIBERO; all variants use LIBERO as action data)\\[4pt]
  \small\setlength{\tabcolsep}{4pt}
  \begin{tabular}{llrrrrrr}
    \toprule
    Init & Training mix & Avg. & Move & Plan & Subtask & BBox & Grip \\
    \midrule
    PaliGemma-ETC & Action only  &  0.05 &  0.18 &   0.00 &  0.08 &  0.00 &  0.00 \\
    PaliGemma-ETC & $+$ MM+ETC   & 72.46 & 43.76 & 100.00 & 54.38 & 75.28 & 88.88 \\
    PaliGemma-ETC & $+$ ETC      & 60.00 & 41.66 & 100.00 & 52.78 & 32.61 & 72.97 \\
    PaliGemma     & Action only  &  0.05 &  0.19 &   0.00 &  0.08 &  0.00 &  0.00 \\
    PaliGemma     & $+$ ETC      & 58.16 & 40.48 & 100.00 & 51.02 & 27.69 & 71.59 \\
    PaliGemma     & $+$ MM+ETC   & 72.30 & 41.87 & 100.00 & 53.65 & 76.72 & 89.28 \\
    \bottomrule
  \end{tabular}

  \vspace{8pt}

  % ---- MM ----
  \phantomsection\label{tab:midtrain_mm_details}%
  \textbf{General MM benchmark}\\[4pt]
  \small\setlength{\tabcolsep}{4pt}
  \begin{tabular}{lllrrrr}
    \toprule
    Init & Training mix & Action data & Avg. & A-OKVQA & COCO & LaTeX-OCR \\
    \midrule
    PaliGemma-ETC & Action only  & Bridge &  0.89 &  0.00 &  0.00 &  2.66 \\
    PaliGemma-ETC & $+$ MM       & Bridge & 43.76 & 51.56 & 13.70 & 66.03 \\
    PaliGemma-ETC & $+$ MM+ETC   & Bridge &  1.09 &  0.00 &  0.00 &  3.27 \\
    PaliGemma     & Action only  & Bridge & 23.65 &  0.03 &  9.02 & 61.91 \\
    PaliGemma     & $+$ MM       & Bridge & 31.11 & 32.49 & 12.25 & 48.58 \\
    PaliGemma     & $+$ MM+ETC   & Bridge & 19.20 &  2.79 &  8.56 & 46.25 \\
    \midrule
    PaliGemma-ETC & Action only  & LIBERO &  0.00 &  0.00 &  0.00 &  0.00 \\
    PaliGemma-ETC & $+$ MM       & LIBERO & 22.97 & 14.76 &  9.39 & 44.77 \\
    PaliGemma-ETC & $+$ MM+ETC   & LIBERO & 17.62 &  0.00 &  8.37 & 44.48 \\
    PaliGemma     & Action only  & LIBERO &  0.00 &  0.00 &  0.00 &  0.00 \\
    PaliGemma     & $+$ MM       & LIBERO & 23.28 & 14.67 & 10.25 & 44.93 \\
    PaliGemma     & $+$ MM+ETC   & LIBERO & 17.54 &  0.00 &  9.02 & 43.60 \\
    \bottomrule
  \end{tabular}

  \vspace{8pt}

  % ---- COCO diagnostics ----
  \phantomsection\label{tab:coco_detection_diagnostics}%
  \textbf{COCO detection diagnostics}
  (Cls-F1@0.6 = class-aware F1; Box-F1@0.6 = localization only; mIoU = mean IoU over matched boxes; Exact = image-level exact-match accuracy)\\[4pt]
  \scriptsize\setlength{\tabcolsep}{2.5pt}
  \begin{tabular}{lllrrrr}
    \toprule
    Init & Training mix & Action data & Cls-F1@0.6 & Box-F1@0.6 & mIoU & Exact \\
    \midrule
    PaliGemma-ETC & Action only & Bridge &  0.00 &  0.05 &  0.00 & 0.00 \\
    PaliGemma-ETC & $+$ MM      & Bridge & 13.70 & 14.23 & 77.46 & 8.24 \\
    PaliGemma-ETC & $+$ MM+ETC  & Bridge &  0.00 &  0.04 &  0.00 & 0.00 \\
    PaliGemma     & Action only & Bridge &  9.02 & 10.06 & 76.24 & 5.09 \\
    PaliGemma     & $+$ MM      & Bridge & 12.25 & 12.92 & 75.89 & 6.36 \\
    PaliGemma     & $+$ MM+ETC  & Bridge &  8.56 &  9.47 & 75.49 & 4.81 \\
    \midrule
    PaliGemma-ETC & Action only & LIBERO &  0.00 &  0.06 &  0.00 & 0.00 \\
    PaliGemma-ETC & $+$ MM      & LIBERO &  9.39 & 10.28 & 75.43 & 5.13 \\
    PaliGemma-ETC & $+$ MM+ETC  & LIBERO &  8.37 &  9.30 & 75.23 & 4.60 \\
    PaliGemma     & Action only & LIBERO &  0.00 &  0.04 &  0.00 & 0.00 \\
    PaliGemma     & $+$ MM      & LIBERO & 10.25 & 11.29 & 75.73 & 5.47 \\
    PaliGemma     & $+$ MM+ETC  & LIBERO &  9.02 & 10.02 & 75.28 & 4.34 \\
    \bottomrule
  \end{tabular}
\end{table*}

\section{Real-World Experiments}
\label{app:real_world_experiments}
% \subsection{Robot Platforms}
% \label{app:real_world_platforms}

% \noindent\textbf{WidowX Platform.}
% We conduct real-world validation on a WidowX 250S with a fixed third-person RealSense camera following the BridgeData V2 setup. Policies take a single RGB observation, proprioceptive state, and a language instruction as input, and predict 10-step action chunks of end-effector delta commands and gripper actions.

% \noindent\textbf{Franka Platform.}
% We also instantiate the protocol on a Franka setup following the DROID data-collection convention, with one Franka arm, two external RGB cameras, and one wrist camera. Policies use the same action interface as WidowX, conditioned on multi-view RGB observations, proprioceptive state, and a language instruction.

\subsection{Task Suite and Deployment Splits}
\label{app:real_world_task_suite}
We evaluate real-world adaptation on two WidowX tasks following the Bridge dataset. Each task has an in-domain split used for task-specific Retentive Adaptation and an OOD split that changes task-relevant visual-language factors without adding OOD action demonstrations. All real-world experiments use the same training hyperparameters as the simulation experiments, except that we set the maximum number of training steps to 30{,}000.

\noindent\textbf{WidowX Tasks.}
The coffee-bag task requires selecting a handwritten-labeled coffee-bean bag and placing it onto a specified plate, e.g., \emph{``Put the ETH coffee bag on the blue plate.''} During task-specific Retentive Adaptation, the model only sees a fixed set of bag labels and plate colors; the OOD split changes both factors, requiring an unseen label--plate color combination.
The flower task requires inserting an instructed flower into a green or white vase, e.g., \emph{``Put the monstera in the green vase.''} Its Retentive Adaptation split contains seen flower--vase pairs, while the OOD split replaces a seen flower category with an unseen one. Thus, the OOD factors in both tasks are held out from task-specific Retentive Adaptation.

% \noindent\textbf{Franka Task Following DROID.}
% The BBQ task requires selecting an instructed sauce or seasoning and applying it to a specified skewer. Its OOD split introduces unseen food and seasoning factors, testing compositional generalization under the DROID-style multi-camera setup.

% \begin{table}[h]
% \centering
% \caption{Real-world task splits. OOD factors are absent from task-specific Retentive Adaptation.}
% \label{tab:realworld_task_splits}
% \small
% \setlength{\tabcolsep}{6pt}
% \begin{tabular}{lll}
% \toprule
% Task & In-domain split & OOD split \\
% \midrule
% Coffee bag & Seen labels and plates & Unseen label + plate color \\
% Flower & Seen flower-vase pairs & Unseen flower category \\ \bottomrule
% \end{tabular}
% \end{table}

\subsection{Real-World Data, Supervision, and Evaluation Protocol}
\label{app:real_world_data_eval}

\begin{figure}
    \centering
    \includegraphics[width=1\linewidth]{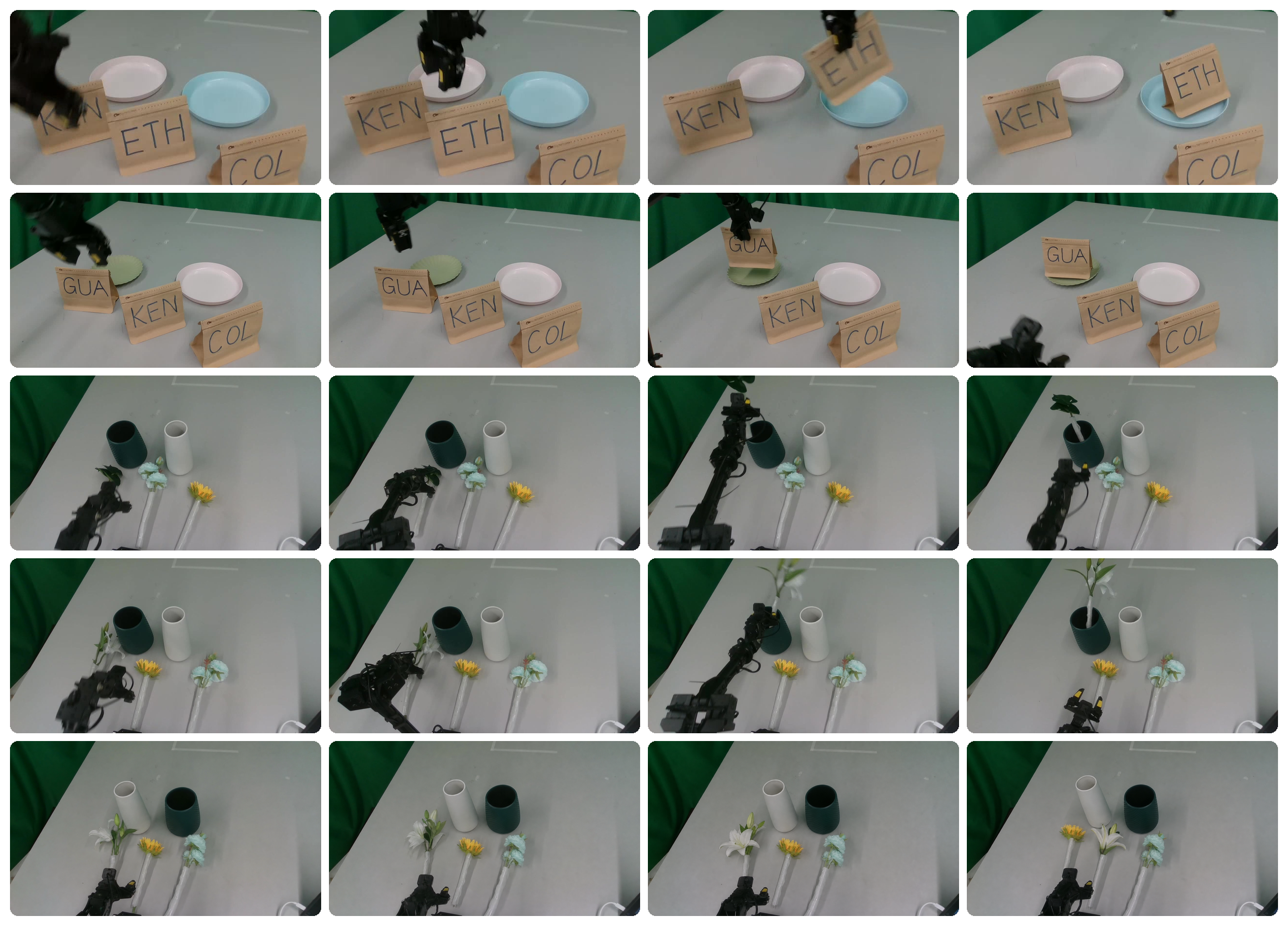}
    \caption{
Real-world task and data examples used in the WidowX experiments.
Rows show, from top to bottom: coffee-bag in-domain rollouts, coffee-bag OOD rollouts, flower in-domain rollouts, flower OOD rollouts, and GPT-Image-2 synthesized flower OOD examples.}
    \label{fig:real_world_data_example}
\end{figure}

Figure~\ref{fig:real_world_data_example} summarizes the real-world ETC data sources used in our Retentive Adaptation ablations. For coffee-bag placement, we separately train with human ETC, computed from real-robot trajectories, and Gemini ETC, generated from the first trajectory frame, to test whether automatically generated annotations can match human trajectory-derived supervision. For flower insertion, we additionally compare real OOD data with GPT-Image-2 synthesized OOD images, evaluating whether generated visual variations can support scalable OOD ETC construction.

\begin{table}[h]
\centering
\caption{Real-world ETC data statistics.}
\label{tab:realworld_etc_stats}
\small
\setlength{\tabcolsep}{5pt}
\begin{tabular}{llcc}
\toprule
Task & ETC source & Records & Images \\
\midrule
\multirow{4}{*}{Coffee bag}
  & ID ETC (Human) & 2880 & 2160 \\
  & ID ETC (Gemini) & 3960 & 2520 \\
  & OOD ETC (Human) & 1120 & 840 \\
  & OOD ETC (Gemini) & 1540 & 980 \\

\multirow{3}{*}{Flower}
  & ID ETC (Gemini) & 1635 & 872 \\
  & OOD ETC (Human) & 540 & 288 \\
  & OOD ETC (GPT-Image-2)& 470 & 188 \\ \bottomrule
\end{tabular}
\end{table}

Table~\ref{tab:realworld_etc_stats} summarizes the ETC data used in our real-world experiments. We distinguish different ETC sources and train them separately in the corresponding ablations, with the resulting policy performance reported in Table~\ref{tab:realrobot_compositional_induction}. Here, \emph{human} denotes ETC supervision directly computed from real-robot trajectories, while \emph{Gemini} denotes annotations generated by prompting Gemini with the first frame of each real-robot trajectory. For the flower task, we additionally include GPT-Image-2-generated images of OOD scenes to evaluate whether synthetic visual variations can provide useful OOD ETC supervision.

We use the following task-specific scoring method for WidowX evaluation. Scores are averaged over $18$ in-domain and $12$ OOD rollouts for each task. Coffee-bag placement is scored as $1.0$ if the correct bag is placed on the correct plate, $0.25$ if the correct bag and plate are selected but the bag is dropped or not reliably grasped, and $0$ otherwise. Flower insertion is scored as $1.0$ for successful insertion, $0.75$ if the correct flower is grasped but dropped before completion, $0.5$ if the correct flower and vase are selected without insertion, $0.25$ if the correct flower is selected but the wrong vase is targeted, and $0$ otherwise.

\section{How ETC Reshapes Embodied Representations}
\label{app:representation_analysis}

\subsection{Distribution Bridging Shapes VLM Representations}
\label{app:representation_analysis_for_stage1}
\begin{figure}
    \centering
    \includegraphics[width=1\linewidth]{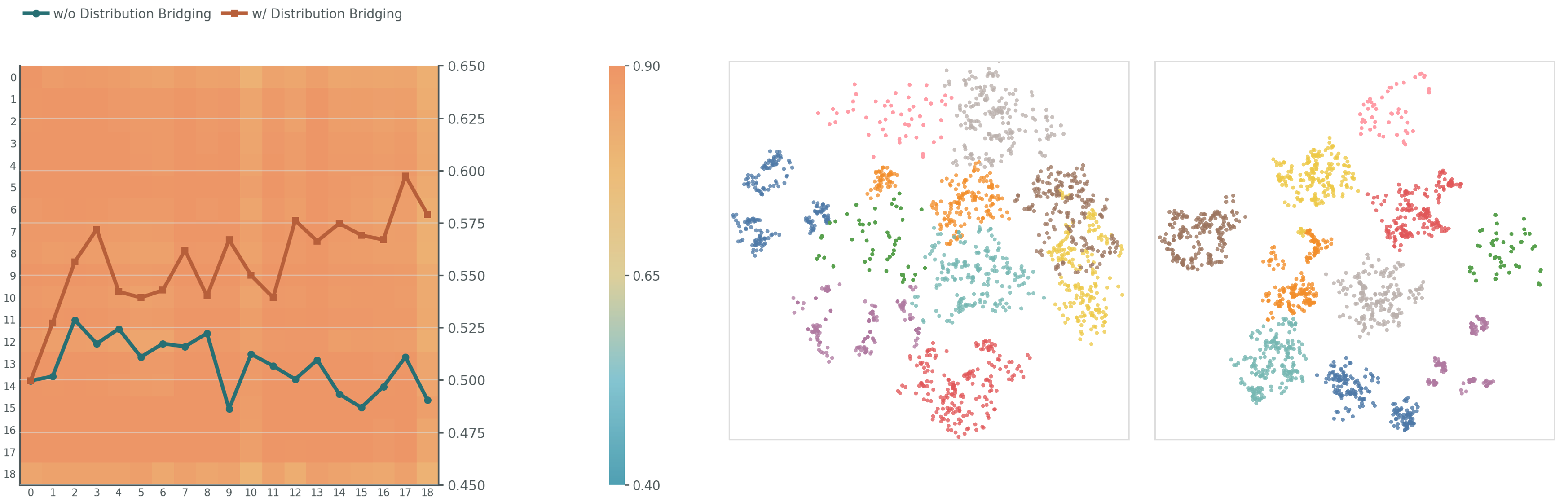}
    \caption{\textbf{Distribution Bridging provides an embodied initialization while preserving the pretrained VLM structure.}
Left: CKA similarity between the original VLM and the ETC-pretrained VLM, together with affordance prediction probing on VLABench.
Right: t-SNE visualizations of VLABench task-conditioned representations before and after Distribution Bridging, where colors denote different task categories.
}
    \label{fig:stage1_representation}
\end{figure}
Figure~\ref{fig:stage1_representation} examines how Distribution Bridging changes the VLM before any action training. The CKA analysis shows that ETC pretraining preserves substantial similarity to the original VLM representation, indicating that embodied adaptation does not overwrite the pretrained visual-language structure. At the same time, layer-wise affordance prediction probing improves after Distribution Bridging, showing that the backbone encodes more manipulation-relevant affordance information.

The t-SNE visualization provides a complementary view of this change. Without Distribution Bridging, representations from different VLABench tasks are more entangled. After Distribution Bridging, task-conditioned embodied inputs form more separable clusters, suggesting that ETC pretraining injects task-discriminative embodied structure into the VLM representation and provides a stronger initialization for downstream policy learning.

\subsection{Preserving Embodied Alignment during Objective Bridging}
\label{app:representation_analysis_for_stage2}
\begin{figure}
    \centering
    \includegraphics[width=1\linewidth]{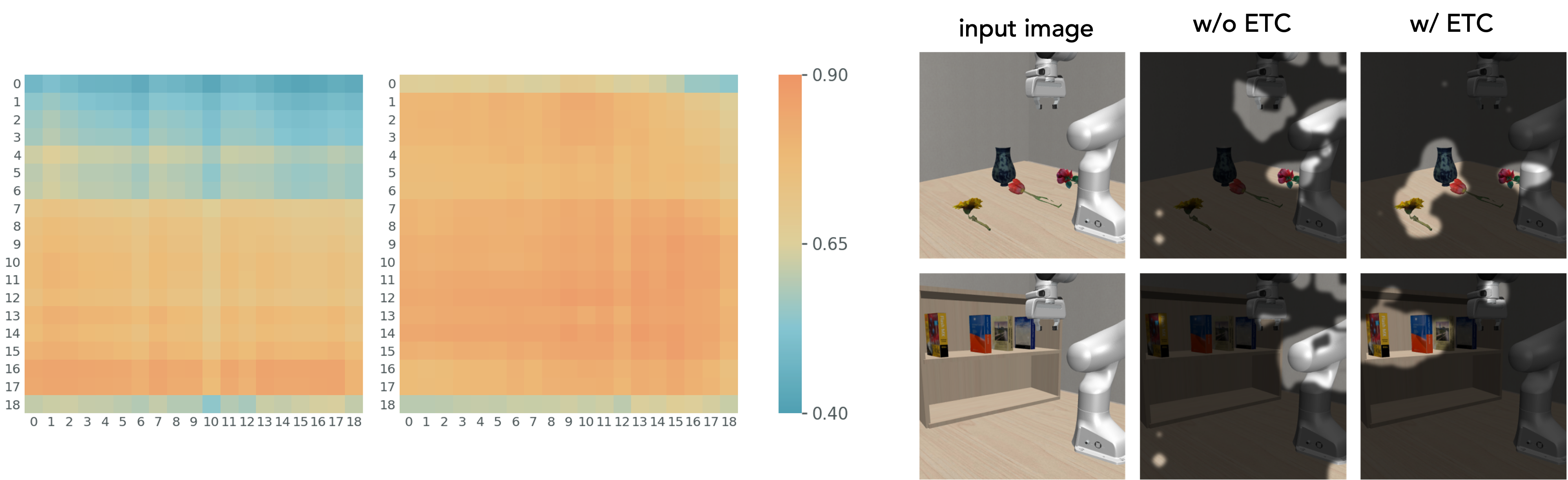}
    \caption{\textbf{Objective Bridging preserves the embodied alignment acquired during Distribution Bridging.}
    Left: CKA similarity between the ETC-pretrained representation and VLA backbones after Objective Bridging with action-only supervision or action--ETC co-training.
    Right: attention visualizations on representative VLABench scenes, comparing the input image, action-only Objective Bridging, and ETC co-training.}
    \label{fig:midtrain_representation}
\end{figure}
Figure~\ref{fig:midtrain_representation} studies whether the embodied alignment acquired from Distribution Bridging is preserved during Objective Bridging. Under action-only Objective Bridging, the policy backbone drifts away from the ETC-pretrained representation, as reflected by lower CKA similarity. In contrast, Objective Bridging with ETC co-training keeps the representation substantially closer to the ETC-pretrained backbone. This suggests that ETC supervision regularizes policy learning and mitigates representational forgetting during action adaptation.

The attention visualizations provide a qualitative view of this preservation effect. Without ETC co-training, the model often attends diffusely or focuses on regions weakly related to the instruction. With ETC co-training, attention is better concentrated on task-relevant objects and interaction regions. Thus, Objective Bridging preserves not only feature-space similarity to the embodied initialization, but also the visual-language grounding needed for downstream policy execution.

\subsection{OOD ETC Induces Compositional Generalization}
\label{app:representation_analysis_for_stage3}
\begin{figure}
    \centering
    \includegraphics[width=1\linewidth]{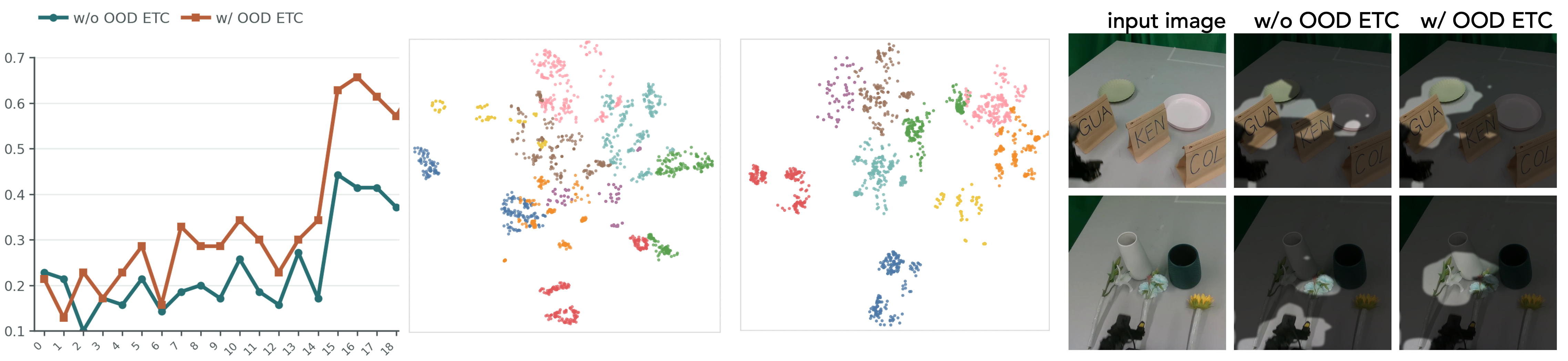}
    \caption{\textbf{OOD ETC induces representations that better support compositional generalization.}
    Left: layer-wise task-planning probing on VLABench Track 2 with and without OOD ETC induction.
Middle: t-SNE visualizations of OOD VLABench Track~2 representations, comparing models trained without and with OOD ETC.
Right: attention visualizations on real-world OOD scenes, comparing the input image, the model without OOD ETC, and the model with OOD ETC.
}
    \label{fig:ood_induce_representation}
\end{figure}
Figure~\ref{fig:ood_induce_representation} examines whether OOD ETC can reshape the policy representation toward unseen deployment conditions. We focus on OOD inputs, where the task-relevant object compositions are absent from action demonstrations. Compared with the model trained without OOD ETC, OOD ETC induction improves layer-wise task-planning probing on OOD embodied inputs, indicating that the backbone better recognizes how unseen objects or compositions should participate in the task. The t-SNE visualization further shows more structured clusters after OOD ETC induction, suggesting that these unseen compositions become more separable in the learned representation.

The attention visualizations make this effect concrete in real-world OOD scenes. Without OOD ETC, attention is often scattered or biased toward irrelevant regions under unseen labels or object categories. With OOD ETC, attention shifts toward the task-relevant objects and interaction regions. These results suggest that OOD ETC provides a lightweight mechanism for binding unseen visual-language factors to existing manipulation skills for compositional generalization.

\end{document}